\def\year{2019}\relax
\documentclass[letterpaper]{article} 
\usepackage{aaai19}  
\usepackage{times}  
\usepackage{helvet}  
\usepackage{courier}  
\usepackage{url}  
\usepackage{graphicx}  
\usepackage{xcolor}
\usepackage{soul}
\usepackage[utf8]{inputenc}
\usepackage[small]{caption}
\usepackage{CJK}
\usepackage{amsmath}
 \usepackage{mathrsfs}
 \usepackage{algorithm}

 \usepackage{subfigure}
\usepackage{algorithmic}
\usepackage{multirow}
\usepackage{xcolor}
\usepackage{setspace}
\usepackage{amsfonts}
\usepackage{colortbl,booktabs}
\usepackage{float}
\begin{document}
%
\title{One-Pass Incomplete Multi-view Clustering}
\author{
$\textbf{Menglei Hu}$,
$\textbf{Songcan Chen}^{*}$
\\
$^1$ College of Computer Science \& Technology, Nanjing University of Aeronautics \& Astronautics \\
$^2$ Collaborative Innovation Center of Novel Software Technology and Industrialization \\
\{ml.hu, s.chen\}@nuaa.edu.cn
}
\maketitle
\begin{abstract}
Real data are often with multiple modalities or from multiple heterogeneous sources, thus forming so-called multi-view data, which receives more and more attentions in machine learning. Multi-view clustering (MVC) becomes its important paradigm. In real-world applications, some views often suffer from instances missing. Clustering on such multi-view datasets is called incomplete multi-view clustering (IMC) and quite challenging. To date, though many approaches have been developed, most of them are offline and have high computational and memory costs especially for large scale datasets. To address this problem, in this paper, we propose an One-Pass Incomplete Multi-view Clustering framework (OPIMC). With the help of regularized matrix factorization and weighted matrix factorization, OPIMC can relatively easily deal with such problem. Different from the existing and sole online IMC method, OPIMC can directly get clustering results and effectively determine the termination of iteration process by introducing two global statistics. Finally, extensive experiments conducted on four real datasets demonstrate the efficiency and effectiveness of the proposed OPIMC method.
\end{abstract}

\section{Introduction}
With the increase of diverse data acquisition devices, real data are often with multiple modalities or from multiple heterogeneous sources \cite{blum1998combining}, forming so-called multi-view data \cite{son2017spectral}. For example, a web document can be represented by its url and words on the page; images of a 3D object are taken from different viewpoints \cite{sun2013survey}. In multi-view datasets, the consistency and complementary information among different views need to be exploited for learning task at hand such as classification and clustering \cite{zhao2017multi}. Nowadays, multi-view learning has been widely studied in different areas such as machine learning, data mining and artificial intelligence \cite{xing2017towards,tulsiani2017multi,nie2018auto}.

Multi-view Clustering (MVC), as one of the most important tasks of multi-view learning, has attracted unimaginable attention due to preventing the expensive requirement of data labeling \cite{bickel2004multi,fan2017robust}. The pursuit of MVC is how to make full use of both consistency and complementary information among multi-view data to get a better clustering result. To date, a variety of related methods have been proposed and these can roughly be divided into two main categories: subspace approaches \cite{Ding2014LowRankCS,cao2015diversity,li2016advances} and spectral approaches \cite{kumar2011co,tao2017ensemble,ren2018robust}. The former try to learn a shared latent subspace such that different dimensionality views are comparable in this space. Whereas the latter aim to learn a unified similarity matrix among multi-view data by extending single-view spectral clustering approaches.

A normal assumption for most of above methods is that all the views are complete, meaning that all the instances appear in individual views and correspond to each other. However, in real-world applications, some views often suffer from instances missing which makes some instances in one view unnecessarily have corresponding instances. Such incompleteness will bring a great difficulty for MVC. Clustering on such incomplete multi-view dataset is called incomplete multi-view clustering (IMC). So far, many approaches have also been developed \cite{li2014partial,shao2015multiple,zhao2016incomplete,Liu2017MultipleKK,wen2018incomplete,hu2018doubly}. Nevertheless, almost all these approaches are offline and can hardly handle large scale datasets because of their high time and space complexities.

In data explosion age, the size of individual views data is often huge. For example, video of hundreds of hours is uploaded to YouTube every minute, which appears in multiple modalities or views, namely audio, text and visual views. Another example is in Web scale data mining, one may encounter billions of Web pages and the dimension of the features may be as large as $\mathcal{O}(10^6)$. Data in such scale is hard to store in the memory and process in offline way. To our best knowledge, to date, only one method OMVC is proposed for the large scale IMC problem \cite{Shao2016OnlineMC}. However, OMVC still suffers from some problems in such aspects as normalizing data matrix, handling missing instances, determining convergence and so on. Therefore, solving large scale IMC problem is still very urgent.

In this paper, we propose an One-Pass Incomplete Multi-view Clustering framework (OPIMC) for large scale multi-view datasets based on subspace learning. OPIMC can easily address IMC problem with the help of Regularized Matrix Factorization (RMF) \cite{Gunasekar2017ImplicitRI,Qi2017Unsupervised} and Weighted Matrix Factorization (WMF) \cite{Kim2009WeightedNM}. Furthermore, OPIMC can directly get clustering results and effectively determine the termination of iteration by introducing the two global statistics which can yield a prominent reduction in clustering time.

In the following, we firstly give a brief review of some related work. Secondly, we detail our OPIMC approach and give the optimization. Thirdly, we report the experimental results. And finally, we conclude the paper.
\section{Related Work}
$\textbf {Multi-view Clustering}$. As mentioned in the introduction, a variety of multi-view clustering methods have been proposed and these can roughly be divided into two categories: subspace approaches \cite{li2016advances} and spectral approaches \cite{ren2018robust}. Contrasting with the spectral approaches, the subspace approaches have become a main paradigm due to less time and space complexities, they try to learn a latent subspace so that different dimensionality views are close to each other in this space. Among the subspace approaches, nonnegative matrix factorization (NMF)\cite{lee1999learning} has become a dominating technique because it can be conveniently applied for clustering and subsequently many NMF based methods and their variants have been proposed. For examples, \cite{liu2013multi} establishes a joint NMF model for multi-view clustering, which performs NMF for each view and pushes low dimensional representation of each view towards a common consensus. Besides, manifold learning is also considered for multi-view clustering problem. By imposing the manifold regularization on the objective function of NMF for data of individual views \cite{wang2016multi,zong2017multi}, these methods get a relatively better results. Here, just to name a few, for more related works on MVC, please refer to \cite{chao2017survey,sun2013survey}\\
$\textbf {Incomplete Multi-view Clustering}$. Most of these previous studies on multi-view clustering assume that all instances present in all views. However, this assumption is not always to be held in real world applications. For example, in the camera network, for some reasons, such as the camera temporarily fail or be blocked by some objects, making the instance missing. This case will cause the incompleteness of multi-view data. Recently, some incomplete multi-view clustering methods have been proposed. For instance,  \cite{li2014partial} proposes PVC to establish a latent subspace where the instances corresponding to the same object in different views are close to each other, and similar instances in the same view should be well grouped by utilizing instance alignment information. Besides, a method of clustering more than two incomplete views is proposed in \cite{shao2015multiple}(MIC) by firstly filling the missing instances with the average feature values in each incomplete view, then handling the problem with the help of weighted NMF and $L_{2;1}$-Norm regularization \cite{kong2011robust,wu2018manifold}. Moreover, \cite{hu2018doubly} proposes DAIMC, which extends PVC to multi-view case by utilizing instance missing information and aligning the clustering centers among different views simultaneously.\\
$\textbf {Online Incomplete Multi-view Clustering}$. In data explosion age, multi-view data tends to be large scale. However the above approaches for incomplete multi-view are almost all offline and can hardly conduct the large scale datasets due to their high time and space complexities. Online learning, as an efficient strategy to build large-scale learning systems, has attracted much attention during the past years \cite{Nguyen2017LargescaleOK,Wan2018EfficientAO}. As a special case of online learning, one-pass learning (OPL) \cite{Zhu2017NewCA} has the benefit of requiring only one pass over the data and is particularly useful and efficient for streaming data. To our best knowledge, to date, only one method extends MIC to online case and develops so-called OMVC \cite{Shao2016OnlineMC} by combining online learning and incomplete multi-view clustering. Nevertheless, OMVC still suffers from some problems in the following aspects:\\
1. Normalization for dataset: OMVC normalizes the multi-view datasets by summing all elements of the data, which is unreasonable in online learning.\\
2. Imputation for missing instances: Due to the mechanism of online learning, it is difficult to get the average feature values in each incomplete view to fill the missing instances.\\
3. Efficiency: OMVC works by learning a consensus latent feature matrix across all the views and then applies \emph{K-means} on this matrix to get the clustering results, which brings high computational cost when both the instance number and the category number are large.\\
4. Termination determination for iterative convergence: OMVC terminates the iteration process by using all the scanned instances, which is not only unreasonable but also time-consuming and laborious.

Considering these disadvantages of the OMVC, we propose a more general and feasible incomplete multi-view clustering algorithm, which can deal with large-scale incomplete multi-view data efficiently and effectively.

\section{Proposed Approach}
\subsection{Preliminaries}
Given an input data matrix $ \textbf X\in\mathbb{R}^{M \times N} $, where each column of $ \textbf X $ is an instance. Regularized Matrix Factorization (RMF) aims to approximately factorize the data matrix $ \textbf X $ into two matrices $ \textbf U $ and $ \textbf V $ with the \emph{Frobenius} norm regularized constraint for $ \textbf U $, $ \textbf V $. Then we can get the following minimization problem
\begin{equation}
\begin{aligned}
\min\limits_{\textbf U,\textbf V}\  \| \textbf X - \textbf U\textbf V^T \|_{\emph{\scriptsize{F}}}^{2} + \alpha\| \textbf U \|_{\emph{\scriptsize{F}}}^{2} + \alpha\| \textbf V \|_{\emph{\scriptsize{F}}}^{2}
\end{aligned}
\end{equation}
where low-rank regularized factor matrices $ \textbf U\in\mathbb{R}^{M \times K} $ and $ \textbf V\in\mathbb{R}^{N \times K} $, \emph{K} denotes dimension of subspace. $\alpha$ is nonnegative parameter. Obviously, this is a biconvex problem. Thus we can easily get the updating rules to find the locally optimal solution for this problem as follows:\\
Update $ \textbf U $ (while fixing $ \textbf V $) using the rule
\begin{equation}
\begin{aligned}
\textbf U=\textbf X\textbf V(\textbf V^{T}\textbf V + \alpha\textbf I_{K})^{-1}
\end{aligned}
\end{equation}
Update $ \textbf V $ (while fixing $ \textbf U $) using
\begin{equation}
\begin{aligned}
\textbf V=\textbf X^T\textbf U(\textbf U^{T}\textbf U + \alpha\textbf I_{K})^{-1}
\end{aligned}
\end{equation}

Weighted Matrix Factorization (WMF), as one of the most commonly used methods for missing matrix, is widely used for recommender systems \cite{Xue2017DeepMF}. The WMF optimization problem is formulated as:
\begin{equation}
\begin{aligned}
\min\limits_{\textbf U,\textbf V}  \| (\textbf X - \textbf U\textbf V^T)\textbf W\|_{\emph{\scriptsize{F}}}^{2}
\end{aligned}
\end{equation}
where \textbf W contains entries only in $\{0, 1\}$, and $\textbf W_{ij}=0$ when the entry $\textbf X_{ij}$ is missing.

\subsection{One-Pass Incomplete Multi-view Clustering}
Given a set of input incomplete multi-view data matrices $\{\textbf X^{(i)}\in\mathbb{R}^{d_{i} \times N},i = 1,2,\cdots,n_{v}\}$, where $d_{i}$, \emph{N} represent the dimensionality and instance number respectively. In order to describe directly and conveniently, the missing instances of individual views are filled with 0. Here we introduce an indicate matrix $ \textbf M\in\mathbb{R}^{n_{v} \times N} $ for this incomplete multi-view dataset.
\begin{eqnarray}
\textbf M_{vj}=
\begin{cases}
1& \text{if $j$-th instance is in the $v$-th view} \\
0& \text{otherwise}
\end{cases}
\end{eqnarray}
where each row of $ \textbf M $ represents the instance presence or absence for corresponding view. From the matrix $\textbf M$, we can easily get the missing information of individual views and aligned information across different views.

For the $v$-th view, inspired by Regularized Matrix Factorization, we can factorize the data matrix $ \textbf X^{(v)}\in\mathbb{R}^{d_{v} \times N} $ into two matrices $\textbf U^{(v)}$ and $\textbf V^{(v)}$, where $ \textbf U^{(v)}\in\mathbb{R}^{d_{v} \times K} $,
$ \textbf V^{(v)}\in\mathbb{R}^{N \times K} $, and \emph{K} denotes dimension of subspace, equal to the categories of the dataset. Furthermore, in order to avoid the third problem of OMVC, we apply an \emph{1-of-K} coding constraint to $\textbf V^{(v)}$, which causes $\| \textbf V^{(v)} \|_{\emph{\scriptsize{F}}}^{2}=N$. Thus we can get the following model:
\begin{equation}
\begin{aligned}
&\min\limits_{\textbf U,\textbf V}  \| \textbf X^{(v)} - \textbf U^{(v)}\textbf V^{(v)T} \|_{\emph{\scriptsize{F}}}^{2} + \alpha\| \textbf U^{(v)} \|_{\emph{\scriptsize{F}}}^{2}\\
&\textrm {s.t.}\ \ \textbf V^{(v)}_{ik} \in \{0,1\}, \sum_{k=1}^{K}\textbf V^{(v)}_{ik} = 1, \forall i=1,2,\cdots,N
\end{aligned}
\end{equation}

For multi-view dataset, (6) does not consider the consistency information across different views. To address this issue, we assume that different views have distinct matrices $\{\textbf U^{(i)}\}_{i=1}^{n_{v}}$ , but share the same matrix $\textbf V$. Meanwhile, we consider the instance missing information to handle the incompleteness of each view with the help of Weighted Matrix Factorization. Thus, (6) is rewritten as:
\begin{equation}
\begin{aligned}
&\min\limits_{\textbf U,\textbf V} \sum_{v=1}^{n_{v}}\{ \| (\textbf X^{(v)} - \textbf U^{(v)}\textbf V^T)\textbf W^{(v)} \|_{\emph{\scriptsize{F}}}^{2} + \alpha\| \textbf U^{(v)} \|_{\emph{\scriptsize{F}}}^{2}\}\\
&\textrm {s.t.}\ \ \textbf V_{ik} \in \{0,1\}, \sum_{k=1}^{K}\textbf V_{ik} = 1, \forall i=1,2,\cdots,N
\end{aligned}
\end{equation}
where the weighted matrix $ \textbf W^{(v)}\in\mathbb{R}^{N \times N} $ is defined as:
\begin{eqnarray}
\textbf W_{jj}^{(v)}=
\begin{cases}
1& \text{if $j$-th instance is in the $v$-th view} \\
0& \text{otherwise}
\end{cases}
\end{eqnarray}

In real-world applications, the data matrices may be too large to fit into the memory. We propose to solve the above optimization problem in an online fashion with low computational and storage complexities. We assume that the data of each view is get by chunks and whose size is $s$. Thus the objective function can be decomposed as:
\begin{equation}
\begin{aligned}
&\mathcal{J} = \sum_{v=1}^{n_{v}}\{\sum_{t=1}^{\lceil N/s\rceil} \| (\textbf X_{t}^{(v)} - \textbf U^{(v)}\textbf V_{t}^T)\textbf W_{t}^{(v)} \|_{\emph{\scriptsize{F}}}^{2} + \alpha\| \textbf U^{(v)} \|_{\emph{\scriptsize{F}}}^{2}\}\\
&\textrm {s.t.}\ \ \textbf V_{ik} \in \{0,1\}, \sum_{k=1}^{K}\textbf V_{ik} = 1, \forall i=1,2,\cdots,N
\end{aligned}
\end{equation}
where $\textbf X_{t}^{(v)}$ is the $t$-th data chunk in the $v$-th view, $\textbf V_{t}$ is the clustering indicator matrix for the $t$-th data chunk, and $\textbf W_{t}^{(v)}$ is the diagonal weight matrix for the $t$-th data chunk.
\subsection{Optimization}
From (9), we can find that it is biconvex for $\{\textbf U^{(v)}\}$ and $\textbf V_t$ at each time $t$. So we update $\{\textbf U^{(v)}\}$ and $\textbf V_t$ in an alternating way. Firstly, we will give the normalization of the dataset.\\
$\textbf {Normalization:}$ In multi-view data, there are scaling differences among views. In order to reduce these differences and improve the clustering results, the appropriate normalization is necessary. However, due to the mechanism of online learning, it is difficult to normalize the dataset using global information such as mean and variance. In this paper, instead we map all the instances to a hypersphere, \emph{i.e.} $\|\textbf X^{(v)}(:,j)\|^2_2=1$.

Next, we describe the following subproblems for the OPIMC optimization problem.\\
$\textbf {Subproblem\ of}$ $\{\textbf U^{(v)}\}_{v=1}^{n_{v}}$. With  $\textbf V_t$ fixed, for each $\textbf U^{(v)}$, the partial derivation of $\mathcal{J}(\textbf U^{(v)})$ with respect to $\textbf U^{(v)}$ is
\begin{equation}
\begin{aligned}
\frac{\partial \mathcal{J}}{\partial \textbf U^{(v)}}=\sum_{i=1}^{t}\ 2(\textbf U^{(v)}\textbf V_i^T-\textbf X_i^{(v)})\textbf W_i^{(v)}\textbf W_i^{(v)^T}\textbf V_i + 2\alpha \textbf U^{(v)}
\end{aligned}
\end{equation}
From the definition of $\textbf W^{(v)}$, we can see that $\textbf W_i^{(v)}=\textbf W_i^{(v)}\textbf W_i^{(v)^T}$. Meanwhile, due to the zero filling of dataset, let $\partial \mathcal{J}/\partial \textbf U^{(v)}=0$, we get the following updating rule:
\begin{equation}
\begin{aligned}
\textbf U^{(v)}=\sum_{i=1}^{t}\ \textbf X_i^{(v)}\textbf V_i(\sum_{i=1}^{t}\textbf V_i^{T}\textbf W_i^{(v)}\textbf V_i + \alpha\textbf I_{K})^{-1}
\end{aligned}
\end{equation}
Here, for the sake of convenience, we introduce two terms $\textbf R_t^{(v)}$ and $\textbf T_t^{(v)}$ as below:
\begin{equation}
\begin{aligned}
\textbf R_t^{(v)}=\sum_{i=1}^{t}\textbf X_i^{(v)}\textbf V_i\ \ \ \ \ \ \
\textbf T_t^{(v)}=\sum_{i=1}^{t}\textbf V_i^{T}\textbf W_i^{(v)}\textbf V_i
\end{aligned}
\end{equation}
Consequently, (11) can be rewritten as:
\begin{equation}
\begin{aligned}
\textbf U^{(v)}=\textbf R_t^{(v)}(\textbf T_t^{(v)} + \alpha\textbf I_{K})^{-1}
\end{aligned}
\end{equation}
Then, when new chunk coming, the matrices $\textbf R_t^{(v)}$ and $\textbf T_t^{(v)}$ can be updating easily as follows:
\begin{equation}
\begin{aligned}
&\textbf R_{t}^{(v)}=\textbf R_{t-1}^{(v)}+\textbf X_t^{(v)}\textbf V_t\\
&\textbf T_{t}^{(v)}=\textbf T_{t-1}^{(v)}+\textbf V_t^{T}\textbf W_t^{(v)}\textbf V_t
\end{aligned}
\end{equation}
$\textbf {Subproblem\ of}$ $\{\textbf V_t\}$. With $\{\textbf U^{(v)}\}_{v=1}^{n_{v}}$ fixed and inspired by \emph{K-means}, we introduce a matrix $\textbf D\in\mathbb{R}^{s \times K}$ to record the distance between all the instances (the column of $\textbf X_t^{(v)}$) and all the clustering centers (the column of $\{\textbf U^{(v)}\}_{v=1}^{n_{v}}$) among all the views.
\begin{equation}
\begin{aligned}
\textbf D_{ij} = \sum_{v=1}^{n_{v}}\| (\textbf X_{t,i}^{(v)} - \textbf U_j^{(v)})\textbf W_{t,ii}^{(v)} \|_{\emph{\scriptsize{F}}}^{2}
\end{aligned}
\end{equation}
where $\textbf X_{t,i}^{(v)}$ denotes the $i$-th instance of $\textbf X_{t}^{(v)}$ and $\textbf W_{t,ii}^{(v)}$ denotes the entry $(i,i)$ of $\textbf W_{t}^{(v)}$.
Note that the indexes of all the row minimum values in matrix $\textbf D$ represent the clustering indicators of the corresponding instances. Thus, we can get the following updating rule for $\textbf V_t$:
\begin{equation}
\begin{aligned}
&\left[\sim ,\emph{index} \right] = \min(\textbf D,[\ ],2),\\
&\textbf V_t = \emph{full}(\emph{sparse}(1:\emph{s},\emph{index},1,\emph{s},\emph{K},\emph{s})).
\end{aligned}
\end{equation}
where (16) is two \emph{matlab} instructions.

From the above procedure, we have solved the first three problems of OMVC \cite{Shao2016OnlineMC}. In the following we will present the solution to OMVC's fourth problem.
$\textbf {Termination determination for iterative convergence:}$ By unfolding the objective function (9), we can get
\begin{equation}
\begin{aligned}
\mathcal{J} =&\sum_{v=1}^{n_{v}}\{\sum_{t=1}^{\lceil N/s\rceil} \| (\textbf X_{t}^{(v)} - \textbf U^{(v)}\textbf V_{t}^T)\textbf W_{t}^{(v)} \|_{\emph{\scriptsize{F}}}^{2} + \alpha\| \textbf U^{(v)} \|_{\emph{\scriptsize{F}}}^{2}\}\\
=&\sum_{v=1}^{n_{v}}\{\sum_{t=1}^{\lceil N/s\rceil} tr(\textbf X_{t}^{(v)T}\textbf X_{t}^{(v)})-2tr(\textbf U^{(v)T}\textbf X_{t}^{(v)}\textbf V_{t})\\
&+tr(\textbf V_{t}^T\textbf W_{t}^{(v)}\textbf V_{t}\textbf U^{(v)T}\textbf U^{(v)})+\alpha\| \textbf U^{(v)} \|_{\emph{\scriptsize{F}}}^{2}\}\\
=&Nn_v(1-ratio) - \sum_{v=1}^{n_{v}}\{2tr(\textbf U^{(v)T}\textbf R_{N}^{(v)})\\
&+tr(\textbf U^{(v)T}\textbf U^{(v)}\textbf T_{N}^{(v)})+\alpha\| \textbf U^{(v)} \|_{\emph{\scriptsize{F}}}^{2}\}
\end{aligned}
\end{equation}
where \emph{ratio} denotes the incomplete rate of the dataset and $tr$ denotes the matrix trace. From (17), by recording the statistics of $\textbf R$ and $\textbf T$, we can easily get the loss of all the scanned instances. Moreover, the memory space requirement for this operation is very small, \emph{i.e. $\mathcal{O}(d_{v}s)$}.

It is worth noting that for the first initial chunk, because of the random initialization of $\textbf U$,$\textbf V$ and the small size of the chunk, in updating $\textbf U$, some clustering centers are likely to be degraded. In order to prevent this, in the iterative update of the first chunk, we use the chunk average values to fill the degenerative clustering centers. While in the iterative update for other chunks, we use the last corresponding values to fill. The experiment results verify the effectiveness of this operation.

The entire optimization procedure for OPIMC is summarized in Algorithm 1.

\begin{algorithm}[h]
\caption{One-Pass Incomplete Multi-view Clustering}
\label{alg1}
\begin{algorithmic}[1]
\REQUIRE
Data matrices for incomplete views $\{\textbf X^{(v)}\}$, weight matrices $\{\textbf W^{(v)}\}$, parameter $\alpha$, number of clusters $K$.\\
\STATE $\textbf R_0^{(v)}=\textbf 0$, $\textbf T_0^{(v)}=\textbf 0$ for each view $v$.
\FOR{$t = 1:\lceil N/s\rceil$}
\STATE Draw $\{\textbf X_t^{(v)}\}$ for all the views.
\IF{$t=1$}
\STATE Initialize the $\{\textbf U^{(v)}\}$, $\textbf V_t$ with random value.
\ELSE
\STATE Initialize the $\textbf V_t$ according to Eq.(15-16).
\ENDIF
\REPEAT
\FOR{$v = 1:n_v$}
\STATE Update $\textbf U^{(v)}$ according to Eq.(11-13).
\ENDFOR
\STATE Fill the degenerative clustering centers
\STATE Update $\textbf V_t$ according to Eq.(15-16)
\UNTIL{converges}
\STATE Update $\textbf R_t^{(v)}$ and $\textbf T_t^{(v)}$ according to Eq.(14).
\ENDFOR
\STATE Get clustering results according to $\textbf V$.
\RETURN
$\{\textbf U^{(v)}\}$ and clustering results.
\end{algorithmic}
\end{algorithm}
\subsection{Convergence}
The convergence of the OPIMC can be proved by the following theorem.\\
$\textbf {Theorem 1}$  \emph{ The objective function value of Eq.(9) is nonincreasing under the optimization procedure in Algorithm 1}.\\
$\textbf {Proof of Theorem 1}$: As shown in Algorithm 1, the optimization of OPIMC can be divided into two subproblems, each of which is convex w.r.t one variable. Thus, by finding the optimal solution for each subproblem alternatively, our algorithm can at least find a locally optimal solution.
\subsection{Complexity}
\emph{$\textbf{Time Complexity}$}: The computational complexity of OPIMC algorithm is dominated by matrix multiplication and inverse operations. We discuss this problem in two aspects: optimizing $\textbf U^{(v)}$, optimizing $\{\textbf V_t\}$. Here we assume that $K \leq d_{v}$, $s$ and $N$. Thus, the time complexities for updating $\textbf U^{(v)}$ and $\{\textbf V_t\}$ are both $\mathcal{O}(d_vKs)$. Suppose $L, d_{max}$ are the iteration times of the loop and the largest dimensionality of all the views respectively, by considering the chunk number $\lceil N/s\rceil$, we can get the overall computational complexity of $\mathcal{O}(Ln_vd_{max}KN)$. It is worth noting that through experiments we find that OPIMC converges quickly, thus setting $L=20$ is enough.\\
\emph{$\textbf{Space Complexity}$}: The proposed OPIMC algorithm only requires $\mathcal{O}(n_vd_{max}s)$ memory space ($s\ll N$). By recording two global statistics $\textbf R$ and $\textbf T$, OPIMC can easily update $\textbf U$, $\textbf V$ and determinate convergence with the scanned instances.
\section{Experiment}
\subsection{DataSets}
In this paper, we conduct the experiments on four real-world multi-view datasets, which contains two small datasets and
\begin{table*}[h]
\centering
\caption{Statistics of the datasets}
\label{tab:datasets}
\begin{tabular}{c|c|c|c}
\hline
Dataset & Instance & View                                                                                  & Cluster \\ \hline
WebKB\footnotemark[1]   & 1051      & Content(3,000), Anchor text (1,840)                                                    & 2        \\
Digit\footnotemark[2]   & 2000      & Fourier (76), Profile (216), Karhunen-Loeve (64), Pixel (240), Zernike (47)            & 10       \\
Reuters\footnotemark[3] & 111740    & English (21,531), French (24,893), German (34,279), Spanish (15,506), Italian (11,547) & 6        \\
Youtube\footnotemark[4] & 92457     & Vision (512), Audio (2,000), Text (1,000)                                              & 31       \\ \hline
\end{tabular}
\end{table*}
\footnotetext[1]{http://vikas.sindhwani.org/manifoldregularization.html}
\footnotetext[2]{http://archive.ics.uci.edu/ml/datasets/Multiple+Features}
\footnotetext[3]{http://archive.ics.uci.edu/ml/machine-learning-databases/ 00259/}
\footnotetext[4]{https://archive.ics.uci.edu/ml/datasets/YouTube+Multiview\\+Video+Games+Dataset}
two large datasets, where Reuters and Youtube are known to be the largest benchmark datasets used for multi-view clustering experiments currently. The important statistics of these datasets are given in the Table \ref{tab:datasets}.

\subsection{Compared Methods}
We compare OPIMC with several state-of-art methods.\\
$\textbf{OPIMC}$: OPIMC is the proposed one-pass incomplete multi-view clustering method in this paper. We search the parameter $\alpha$ in $\{1e\textnormal{-}4,1e\textnormal{-}3,1e\textnormal{-}2,1e\textnormal{-}1,1e0,1e1,1e2,1e3\}$.\\
$\textbf{IMC}$: As shown in (7), IMC is the offline case of OPIMC.\\
$\textbf{OMVC}$: OMVC is an online incomplete multi-view clustering method proposed in \cite{Shao2016OnlineMC}. To facilitate comparison, we set $\alpha_vs (\beta_vs)$ the same value for all the views. Meanwhile, we select the parameter $\alpha$ within the set of $\{1e\textnormal{-}3,1e\textnormal{-}2,1e\textnormal{-}1,1e0\}$ and select the parameter $\beta$ within $\{1e\textnormal{-}7,1e\textnormal{-}6,1e\textnormal{-}5,1e\textnormal{-}4,1e\textnormal{-}3,1e\textnormal{-}2\}$.\\
$\textbf{MultiNMF}$: MultiNMF is a classic offline method for multi-view clustering proposed in \cite{liu2013multi}. We select the parameter $\alpha$ within $\{1e\textnormal{-}3,1e\textnormal{-}2,1e\textnormal{-}1,1e0\}$.\\
$\textbf{ONMF}$: ONMF is an online document clustering algorithm for single view using NMF \cite{wang2011efficient}. In order to apply ONMF, we simply concatenate all the normalized views together to form a big single view. We compare two versions of ONMF from the original paper. $\textbf{ONMFI}$ is the original algorithm that calculates the exact inverse of Hessian matrix, while $\textbf{ONMFDA}$ uses diagonal approximation for the inverse of Hessian matrix.

\subsection{Setup}
To simulate the incomplete view setting, we randomly remove some instances from each view. On WebKB and Digit datasets, we set the incomplete rate to 0.3 and 0.4 respectively for the experiment. Besides, we set the incomplete rate to 0.4 on Reuters and Youtube datasets. Meanwhile we shuffle the order of the samples to fit the more real online scene. The chunk size \emph{s} for online methods is set to 50 for small datasets and 2000 for large datasets, respectively. Meanwhile, it is worth mentioning that MultiNMF and ONMF can only deal with complete multi-view dataset, in order to the completeness of the experiment, we firstly fill the missing instances in each incomplete view using average feature values.

The normalized mutual information (NMI) and precision (AC) clustering evaluation measures are used in this paper. For online and one-pass methods, in order to more comprehensively compare with OMVC and ONMF, \emph{we also conduct the experiments for 10 passes and report both NMI and AC for different passes}. The experimental results are shown in Figure 1.

\begin{figure*}[t]
      \centering
      \subfigure[AC for 0.3 missing WebKB]{
           \label{plot:ACs for 0.3 missing WebKB}
           \includegraphics[width=0.235\textwidth]{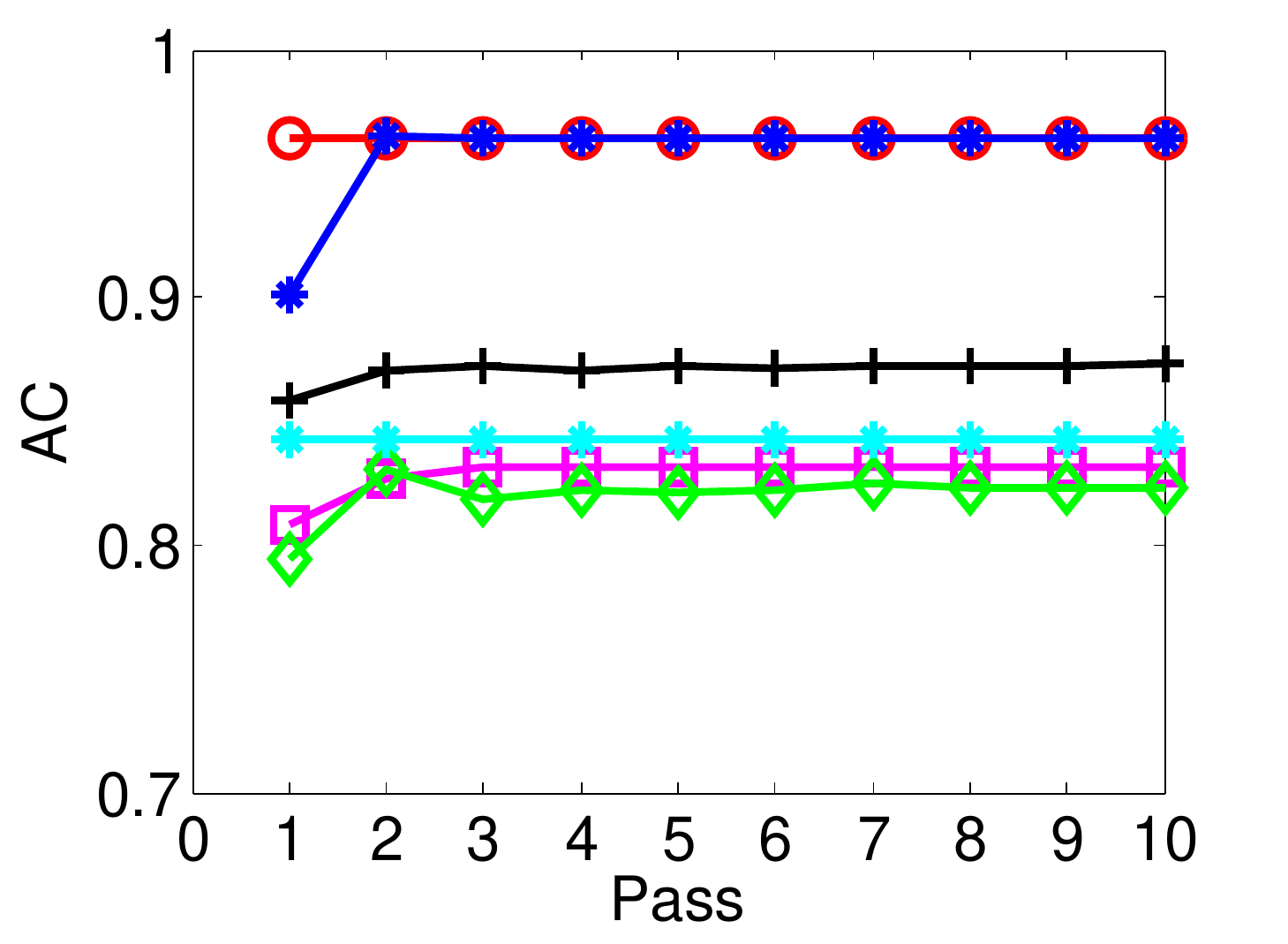}
      }
      \subfigure[NMI for 0.3 missing WebKB]{
           \label{plot:NMIs for 0.3 missing WebKB}
           \includegraphics[width=0.235\textwidth]{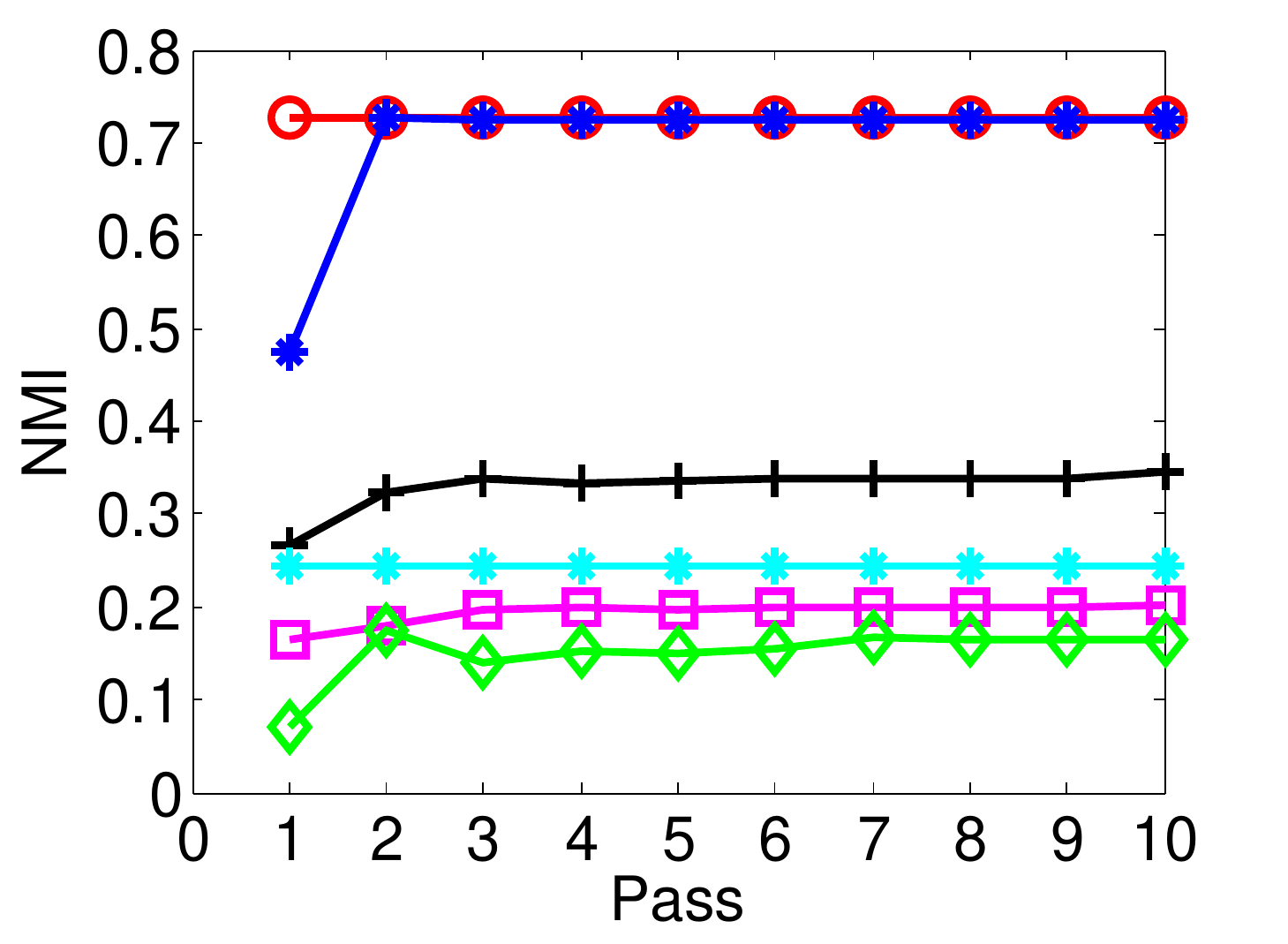}
      }
      \subfigure[AC for 0.3 missing Digit]{
           \label{plot:ACs for 0.3 missing Digit}
           \includegraphics[width=0.235\textwidth]{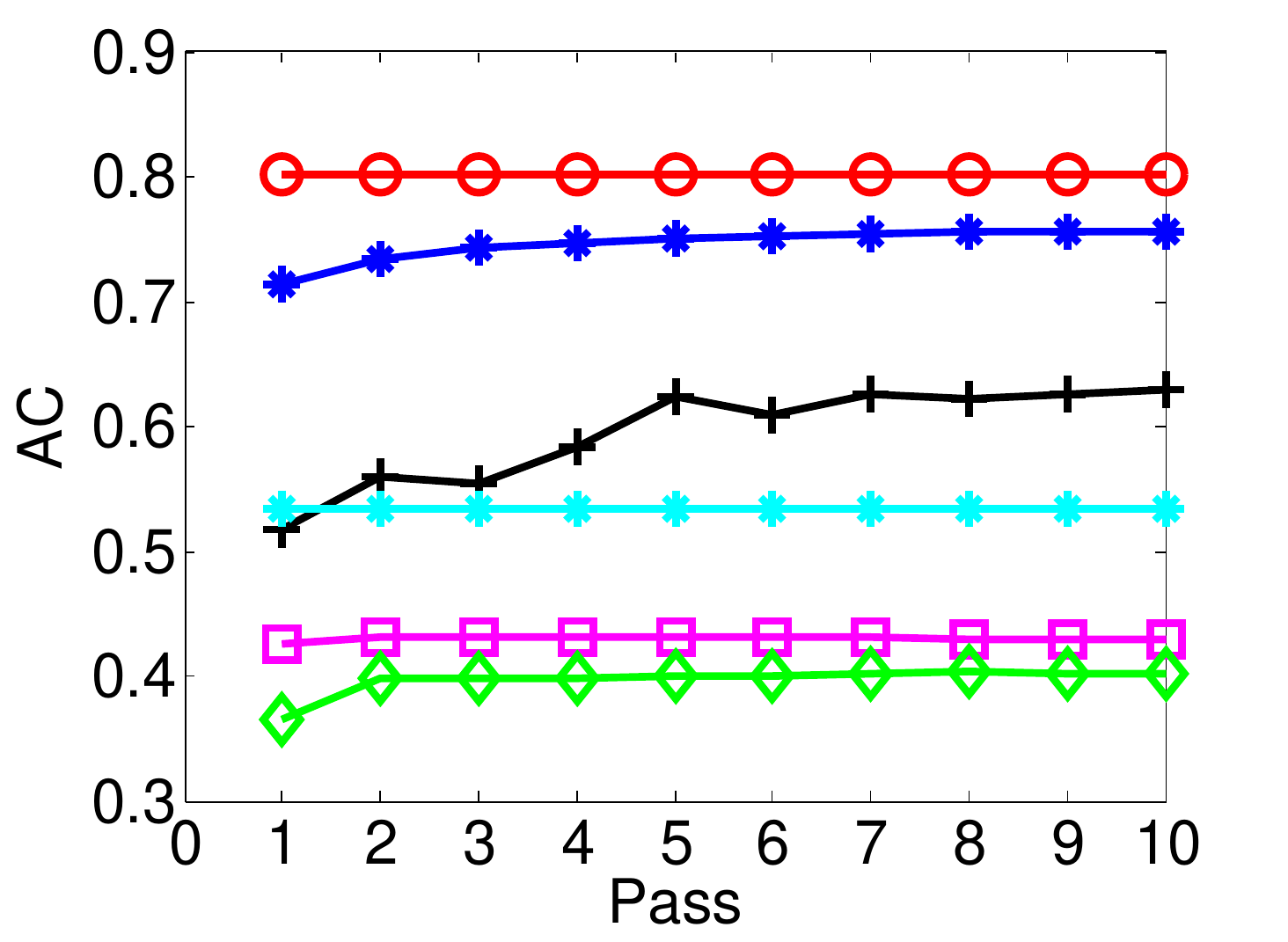}
      }
      \subfigure[NMI for 0.3 missing Digit]{
           \label{plot:NMIs for 0.3 missing Digit}
           \includegraphics[width=0.235\textwidth]{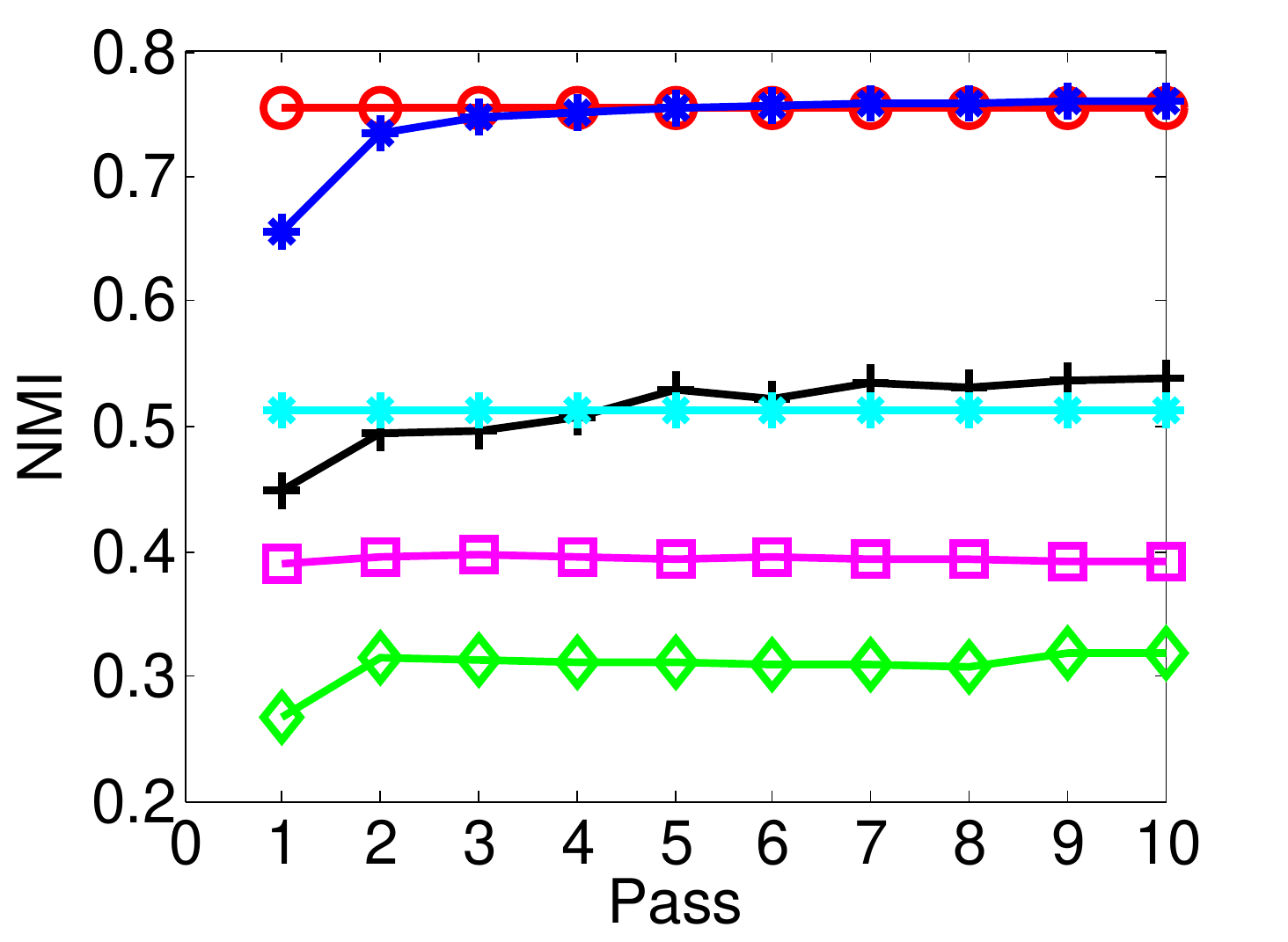}
      }
      \subfigure[AC for 0.4 missing WebKB]{
           \label{plot:ACs for 0.4 missing WebKB}
           \includegraphics[width=0.235\textwidth]{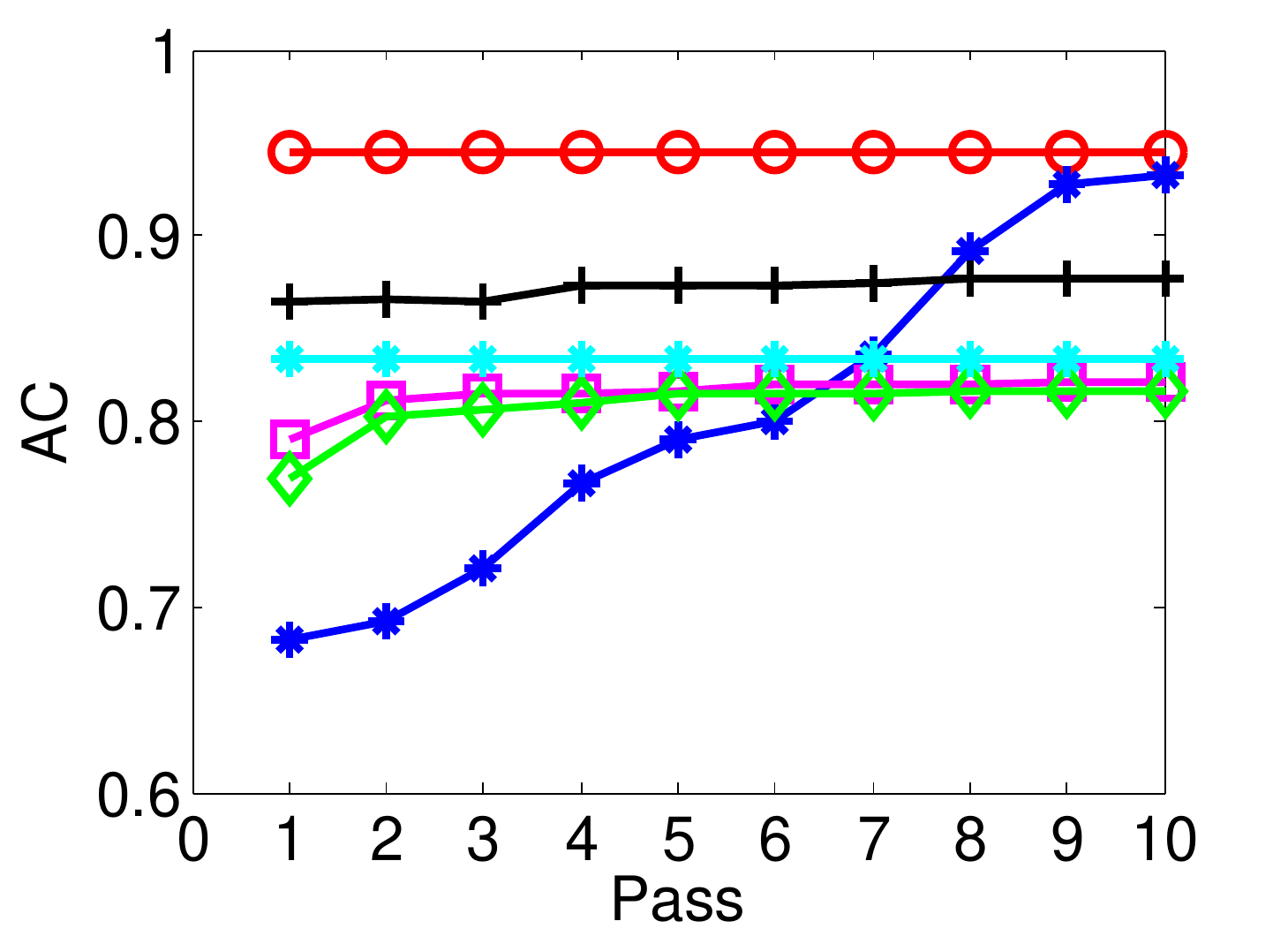}
      }
      \subfigure[NMI for 0.4 missing WebKB]{
           \label{plot:NMIs for 0.4 missing WebKB}
           \includegraphics[width=0.235\textwidth]{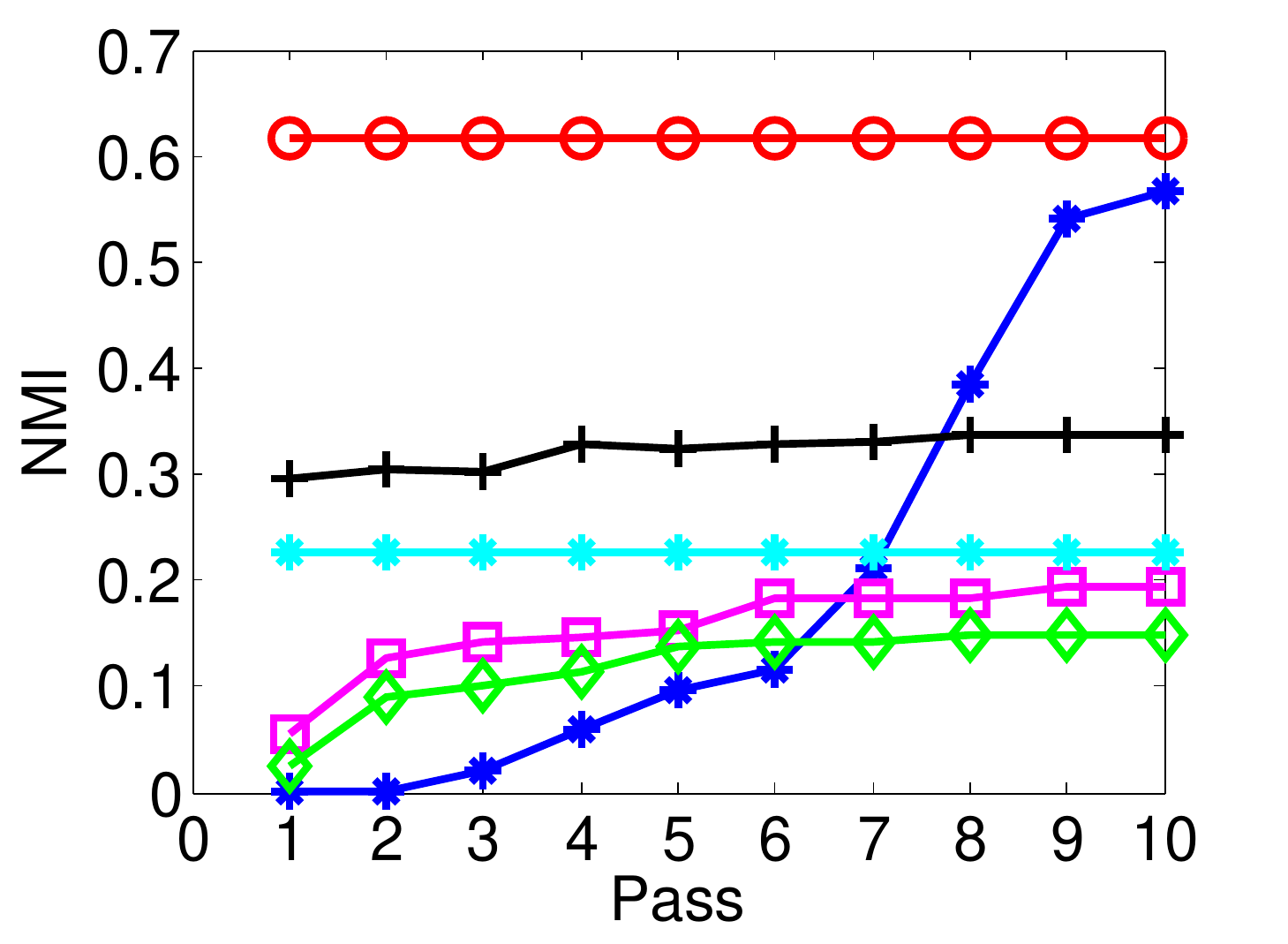}
      }
      \subfigure[AC for 0.4 missing Digit]{
           \label{plot:ACs for 0.4 missing Digit}
           \includegraphics[width=0.235\textwidth]{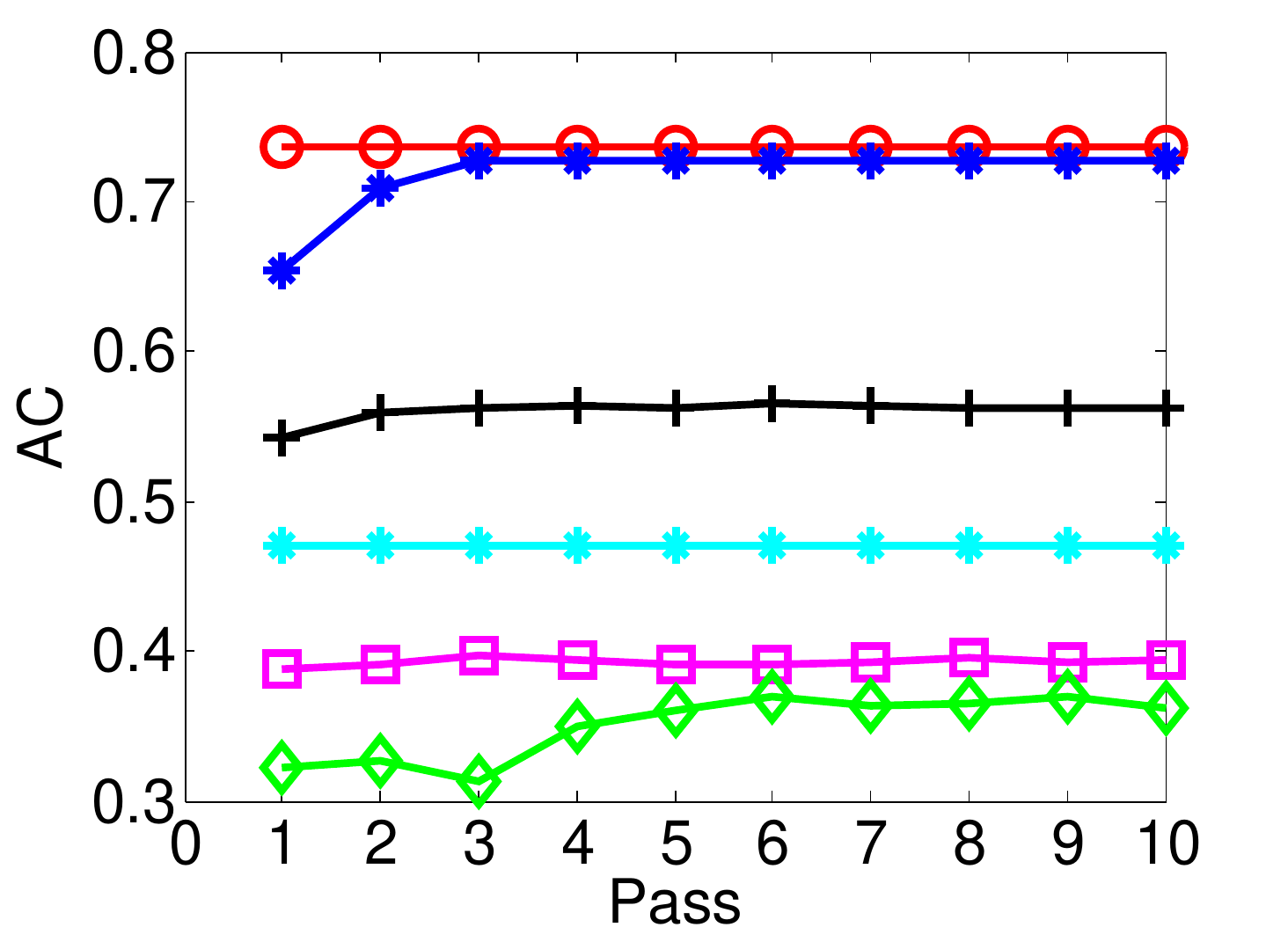}
      }
      \subfigure[NMI for 0.4 missing Digit]{
           \label{plot:NMIs for 0.4 missing Digit}
           \includegraphics[width=0.235\textwidth]{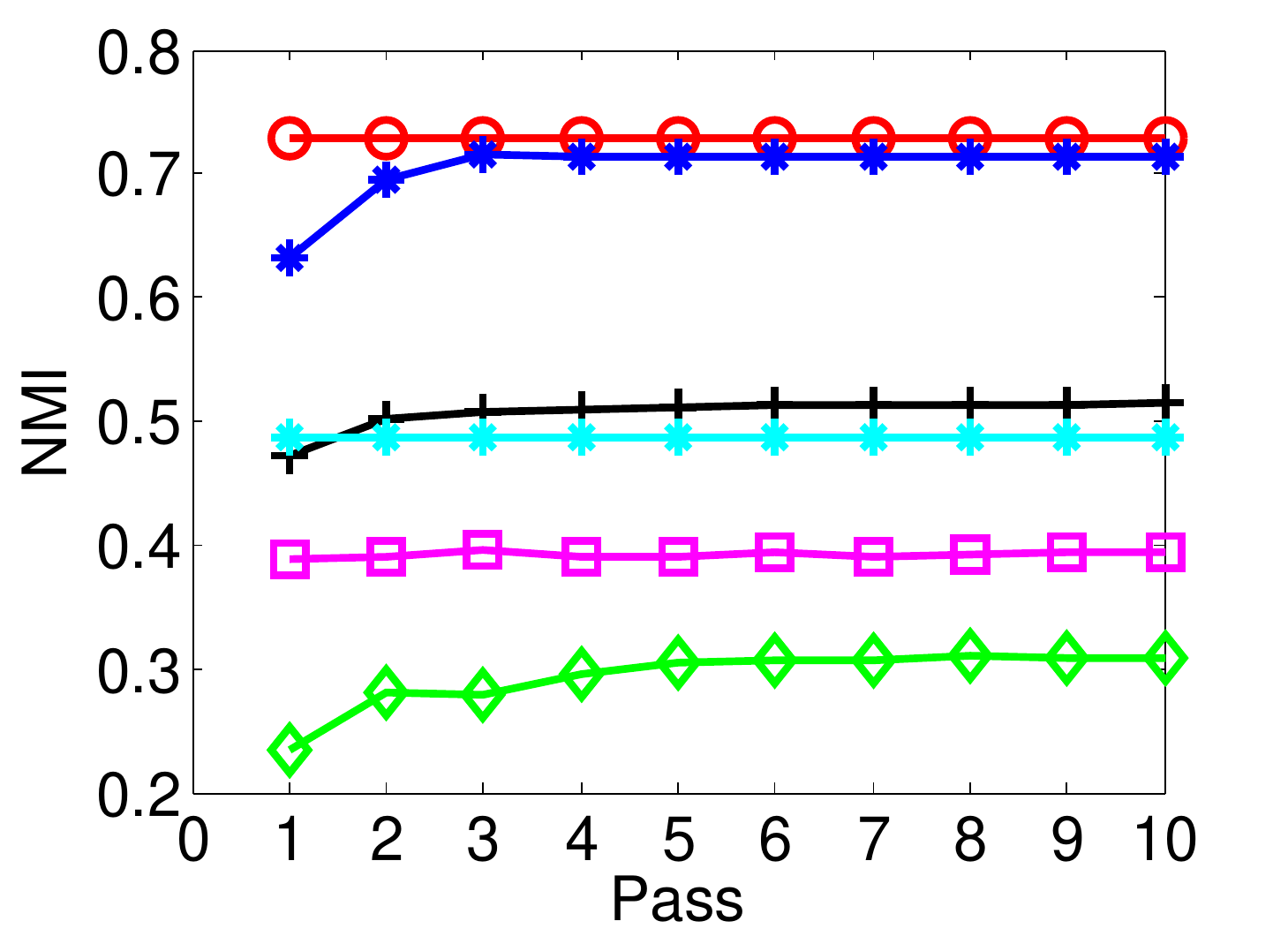}
      }
      \subfigure[AC for 0.4 missing Reuters]{
           \label{plot:ACs for 0.4 missing Reuters}
           \includegraphics[width=0.235\textwidth]{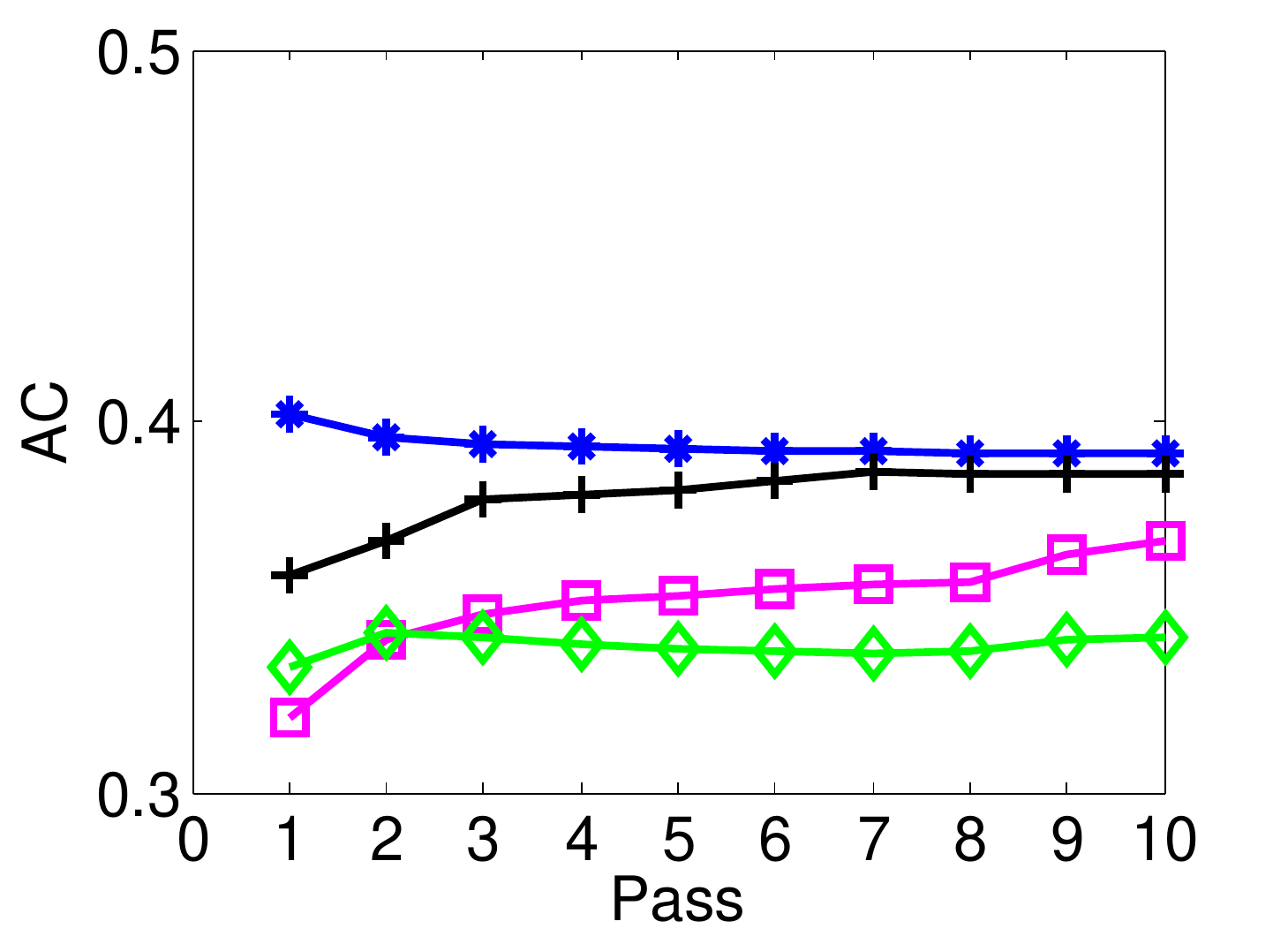}
      }
      \subfigure[NMI for 0.4 missing Reuters]{
           \label{plot:NMIs for 0.4 missing Reuters}
           \includegraphics[width=0.235\textwidth]{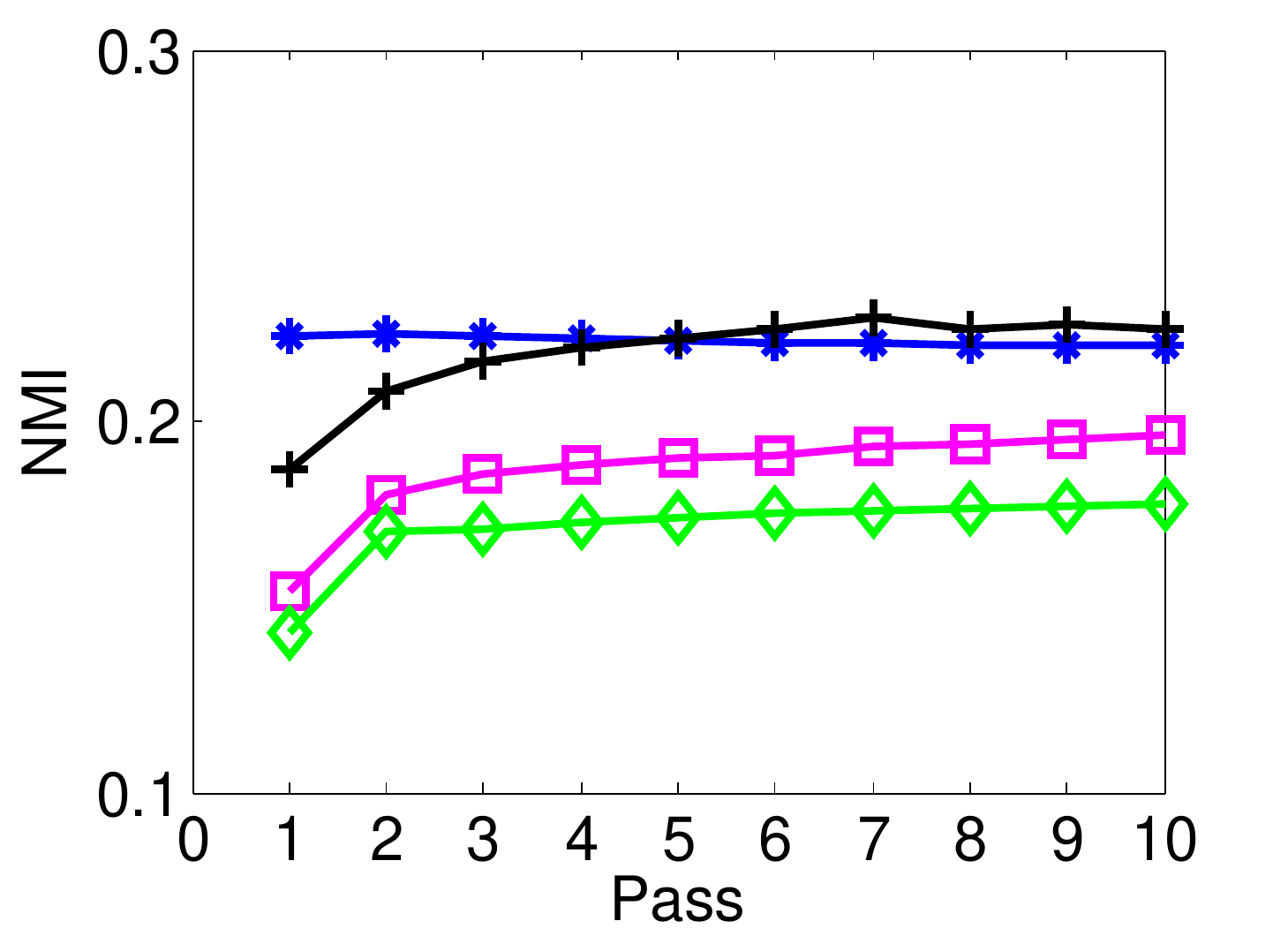}
      }
      \subfigure[AC for 0.4 missing Youtube]{
           \label{plot:ACs for 0.4 missing Youtube}
           \includegraphics[width=0.235\textwidth]{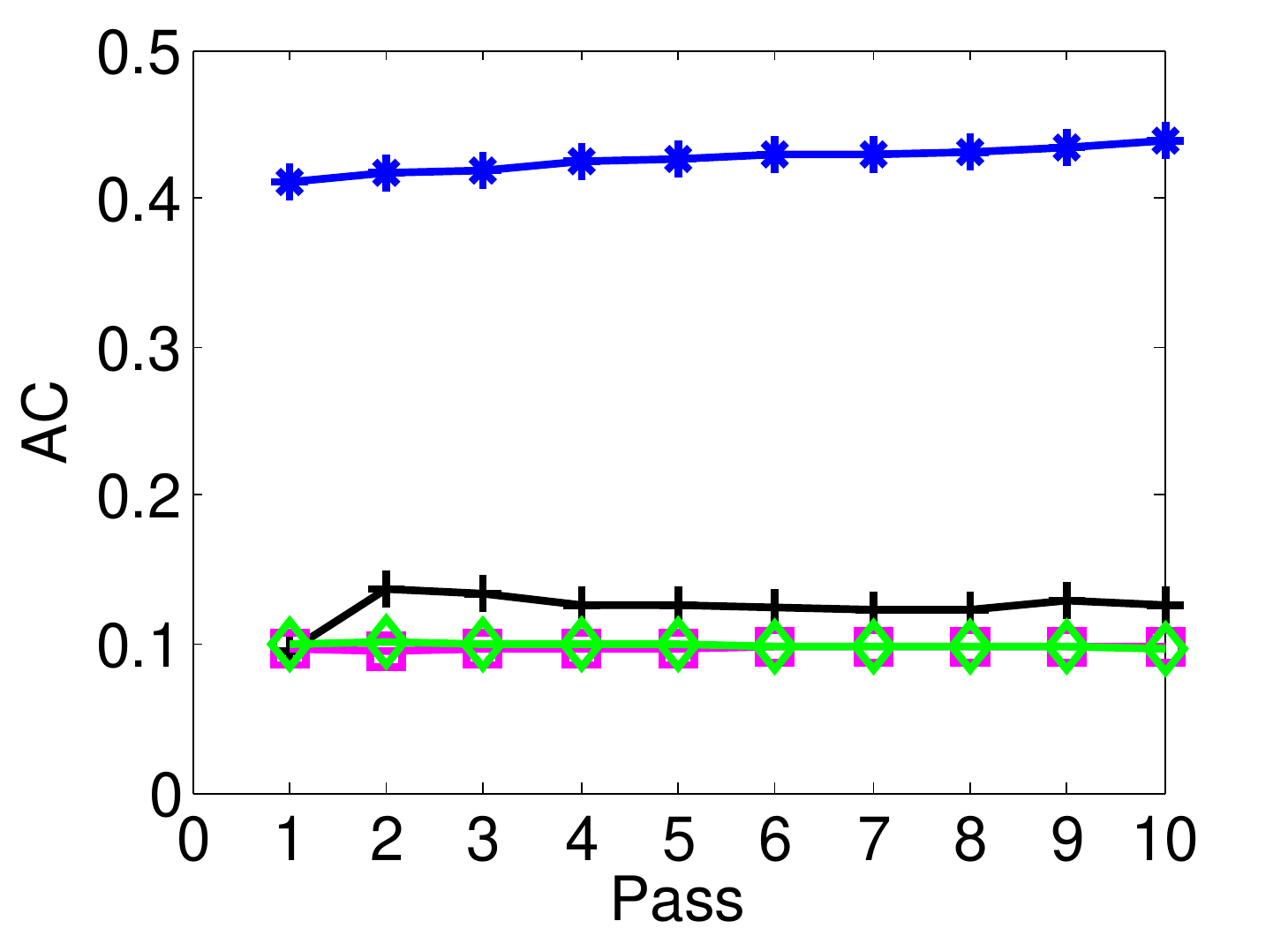}
      }
      \subfigure[NMI for 0.4 missing Youtube]{
           \label{plot:NMIs for 0.4 missing Youtube}
           \includegraphics[width=0.235\textwidth]{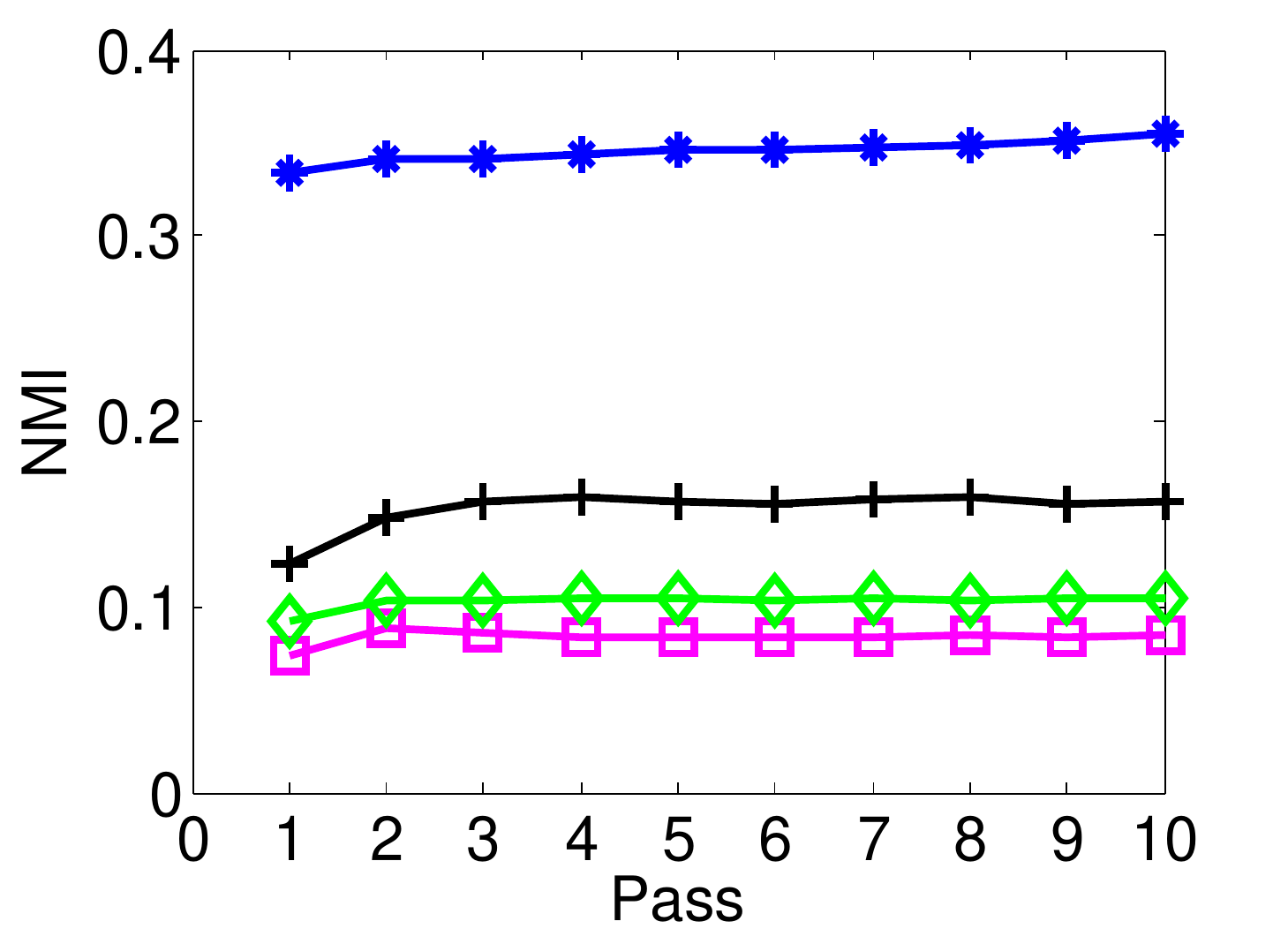}
      }
      \subfigure{
           \label{legend}
           \includegraphics[width=0.5\textwidth]{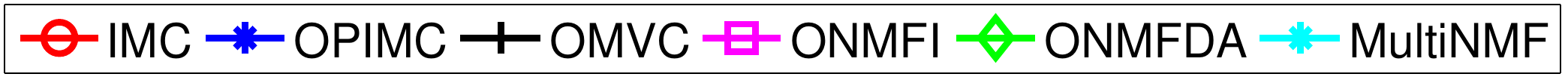}
      }
      \caption{Performance of clustering on WebKB, Digit, Reuters and Youtube for different passes.}
      \label{Fig:NMIAC}
  \end{figure*}
\subsection{Results}
Figure \ref{Fig:NMIAC} reports the performance of clustering on WebKB, Digit, Reuters and Youtube datasets for different passes with different incomplete rates. From Figure \ref{Fig:NMIAC}, we can get the following results.

From Figure \ref{plot:ACs for 0.3 missing WebKB} and Figure \ref{plot:NMIs for 0.3 missing WebKB}, we can see that on WebKB dataset, the offline method IMC achieves the best performance, the proposed OPIMC gets close performance after just two passes and outperforms the other four comparison methods. The same phenomena can be observed from Figures \ref{plot:ACs for 0.3 missing Digit}, \ref{plot:NMIs for 0.3 missing Digit}, \ref{plot:ACs for 0.4 missing Digit} and \ref{plot:NMIs for 0.4 missing Digit} on Digit dataset.

From Figure \ref{plot:ACs for 0.4 missing WebKB} and Figure \ref{plot:NMIs for 0.4 missing WebKB}, we can see that OPIMC performs terribly on WebKB dataset in the first few passes for the incomplete rate of 0.4. The main reasons are that the large incomplete rate and the small size of the chunk, which cause the matrices $\{\textbf U^{(v)}\}$
hard to be learned. However, after few passes, through continuous correction of global information, the clustering performance on WebKB dataset grows rapidly.

On large scale Reuters dataset, from Figure \ref{plot:ACs for 0.4 missing Reuters} and Figure \ref{plot:NMIs for 0.4 missing Reuters}, we can see that OPIMC gets the best results after only one pass, but the clustering performance decreases with the pass number increasing.

From Figure \ref{plot:ACs for 0.4 missing Youtube} and Figure \ref{plot:NMIs for 0.4 missing Youtube}, we can find that on Youtube dataset, OPIMC produces excellent results and much better than the other methods. This fully demonstrates the effectiveness of OPIMC.\\
$\textbf{Complexity\  Study}$: All the experiments are run on computer with Intel(R)390 Core(TM) i5-3470 @ 3.20GHz CPU and 16.0 GB RAM with the help of Matlab R2013a. The complexity study results are reported in Table \ref{tab:time}.
\begin{table}[tbp]
\centering
\caption{Run time for different methods}
\label{tab:time}
\begin{tabular}{c|c|c|c|c}
\hline
            & \multicolumn{4}{c}{Run Time (seconds)}          \\ \hline
            & WebKB & Digit & Reuters & Youtube \\ \hline
OPIMC/Pass  & \textbf{0.25}  & \textbf{0.56}  & \textbf{27.89}   & \textbf{26.76}   \\ \hline
OMVC/Pass   & 23.37 & 34.76 & 3753.02 & 2064.83  \\ \hline
ONMFI/Pass  & 18.69 & 31.16 & 2887.12 & 1657.22  \\ \hline
ONMFDA/Pass & 20.09 & 30.63 & 2224.44 & 1307.14 \\ \hline
IMC         & 2.91   & 6.31  & /       & /       \\ \hline
MultiNMF    & 149.7 & 647.2 & /       & /       \\ \hline
\end{tabular}
\end{table}
\begin{figure}[htb]
      \centering
      \subfigure[]{
           \label{plot:paraAC}
           \includegraphics[width=3.95cm,height=3.0cm]{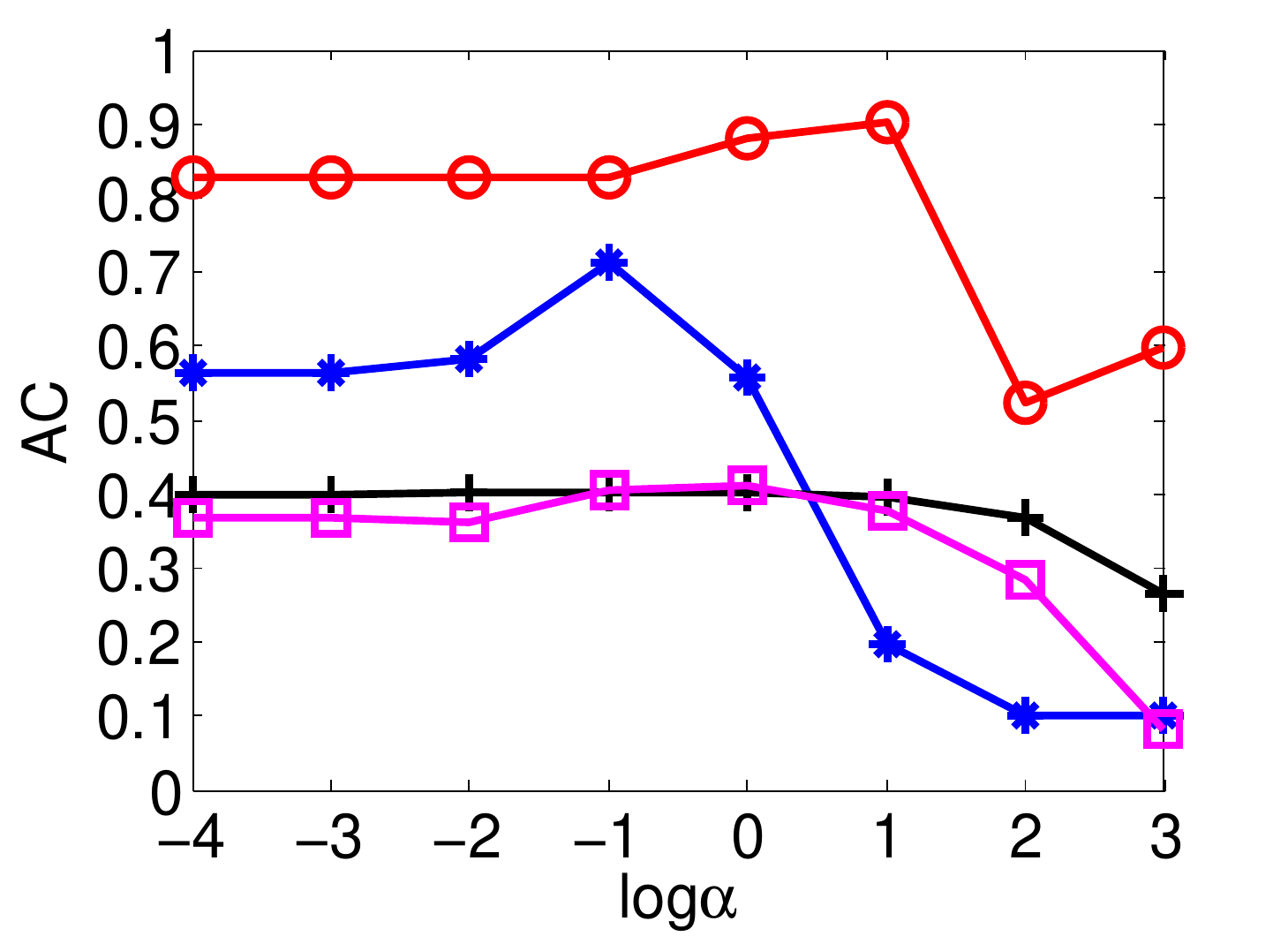}
      }
      \subfigure[]{
           \label{plot:paraNMI}
           \includegraphics[width=3.95cm,height=3.0cm]{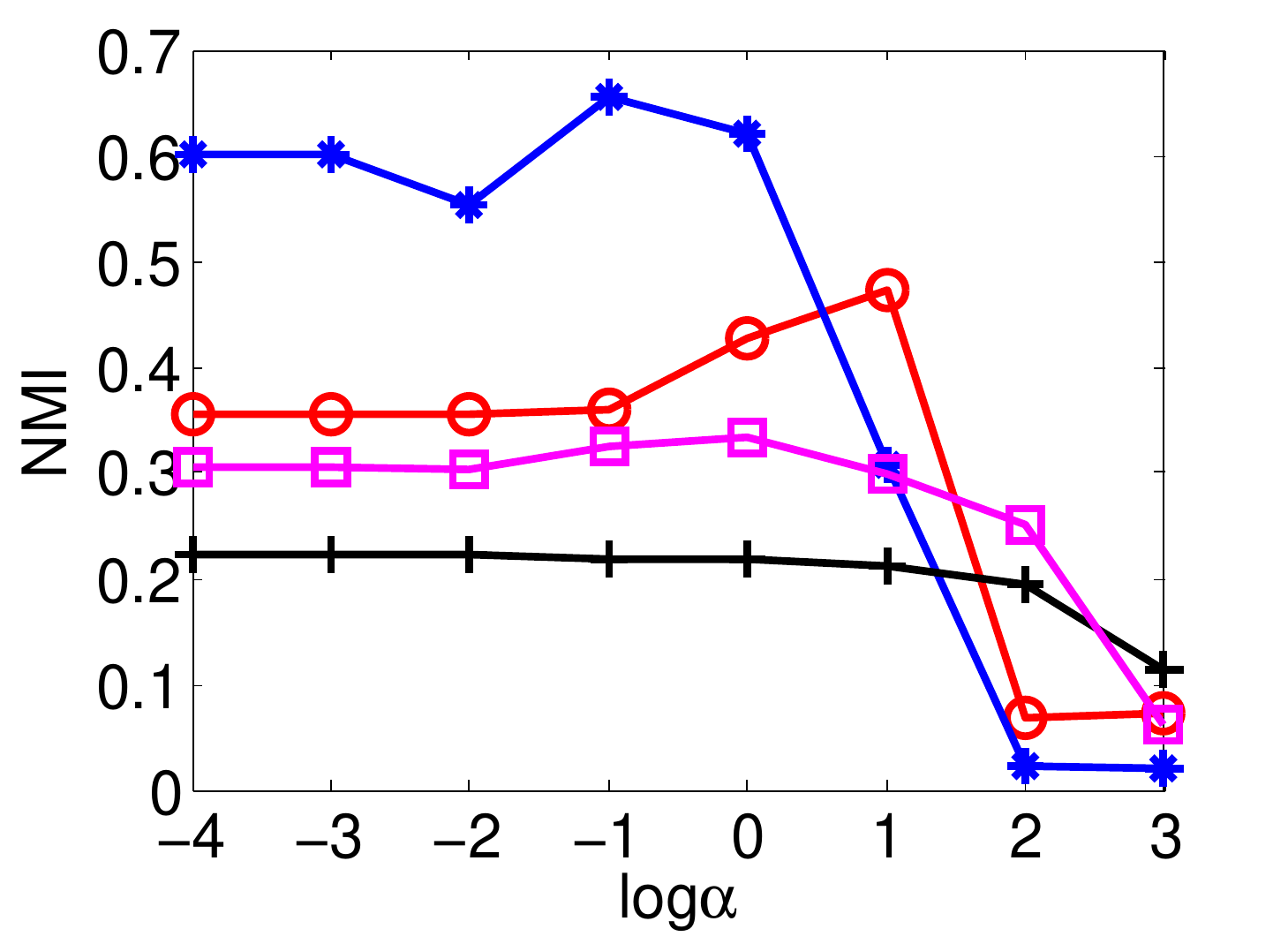}
      }
      \subfigure{
           \label{legend}
           \includegraphics[width=0.35\textwidth]{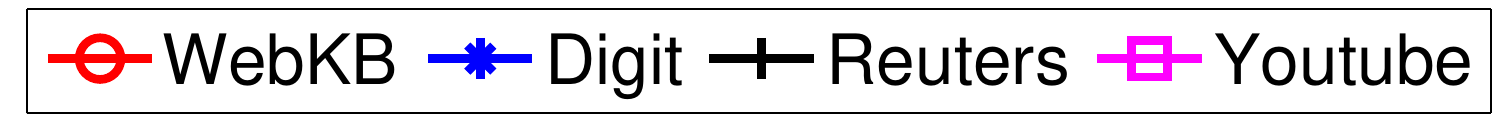}
      }
      \caption{Parameter studies on WebKB, Digit, Reuters and Youtube datasets, where the incomplete rate of WebKB and Digit experiment is set as 0.3, and the incomplete rate of Reuters and Youtube experiment is set as 0.4.}
      \label{Fig:para}
  \end{figure}

From Table \ref{tab:time}, we can get some observations. Firstly, OMVC gets better results than ONMFI and ONMFDA, but the latter two methods run faster than OMVC. Secondly, the offline method IMC runs faster than the other methods except OPIMC. Thirdly, compared with OMVC, OPIMC takes much less running time (only $1\%\textnormal{-}2\%$ of OMVC running time), while obtains relatively better clustering results. All these observations prove the efficiency and effectiveness of our model. \\
$\textbf{Parameter\ Study}$: We conduct the parameter experiments on the four aforementioned datasets for just one pass. Meanwhile, we set the incomplete rate as 0.3 for small datasets and 0.4 for large scale datasets respectively, and report the clustering performance of OPIMC by ranging $\alpha$ in the set of $\{1e\textnormal{-}4,1e\textnormal{-}3,1e\textnormal{-}2,1e\textnormal{-}1,1e0,1e1,1e2,1e3\}$. The results are shown in Figure \ref{Fig:para}.

From Figure \ref{Fig:para}, we can see that OPIMC gets best clustering results in $\alpha =\{1e1,1e\textnormal{-}1,1e\textnormal{-}2,1e0\}$ on WebKB, Digit, Reuters and Youtube datasets respectively.
\\
$\textbf{Convergence\ Study}$: The convergence experiments are conducted on the four aforementioned datasets for 20 passes. We set the incomplete rate as 0.4 for all the datasets and conduct the experiments. According to the definition of $\textbf R^{(v)}$, $\textbf T^{(v)}$, and inspired by ONMF, OMVC, for the first pass, the average loss is defined as follows:
\begin{figure*}[htb]
      \centering
      \subfigure[Average Loss on WebKB]{
           \label{plot:lossWebKB}
           \includegraphics[width=3.95cm,height=3cm]{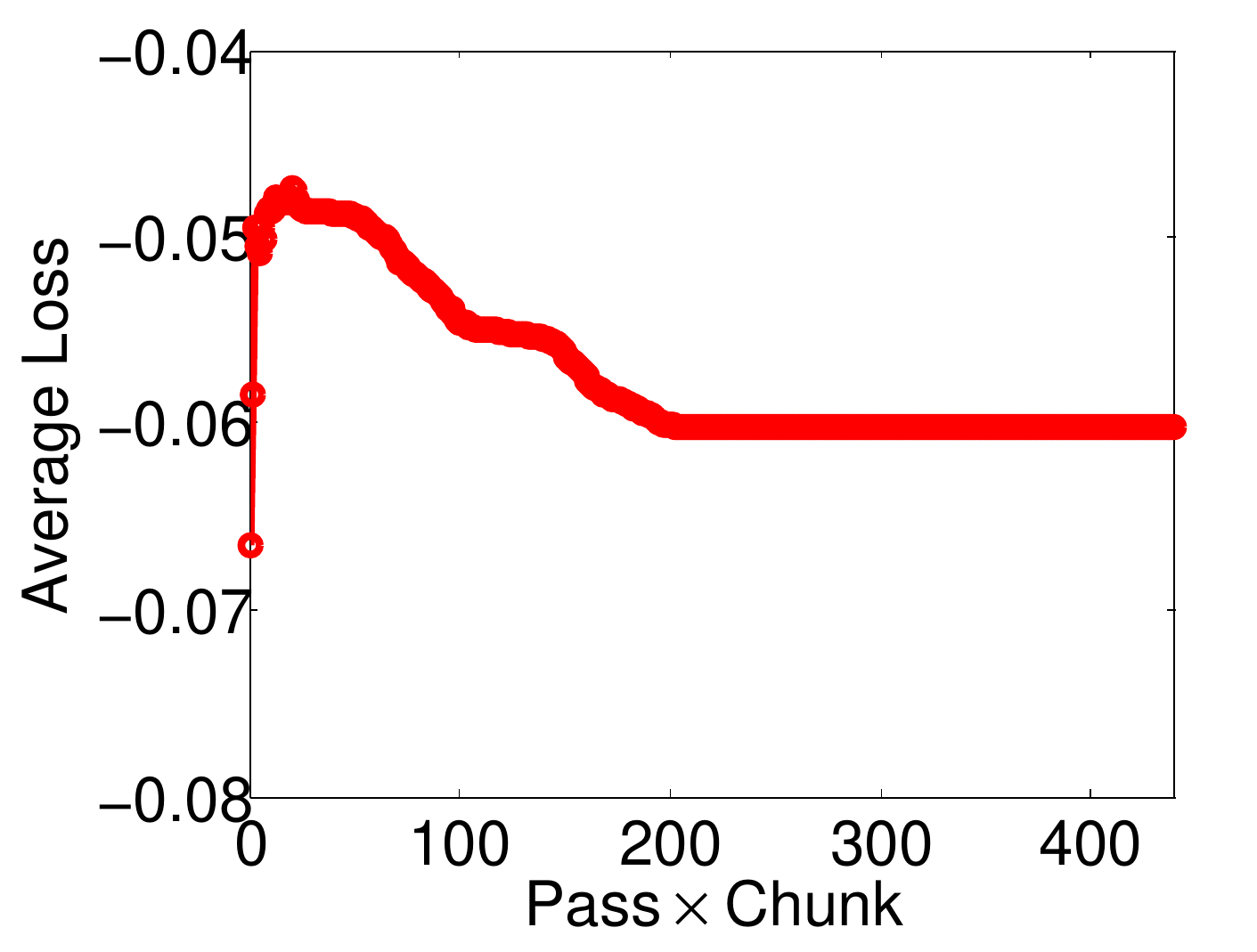}
      }
      \subfigure[Average Loss on Digit]{
           \label{plot:lossDigit}
           \includegraphics[width=4cm,height=3cm]{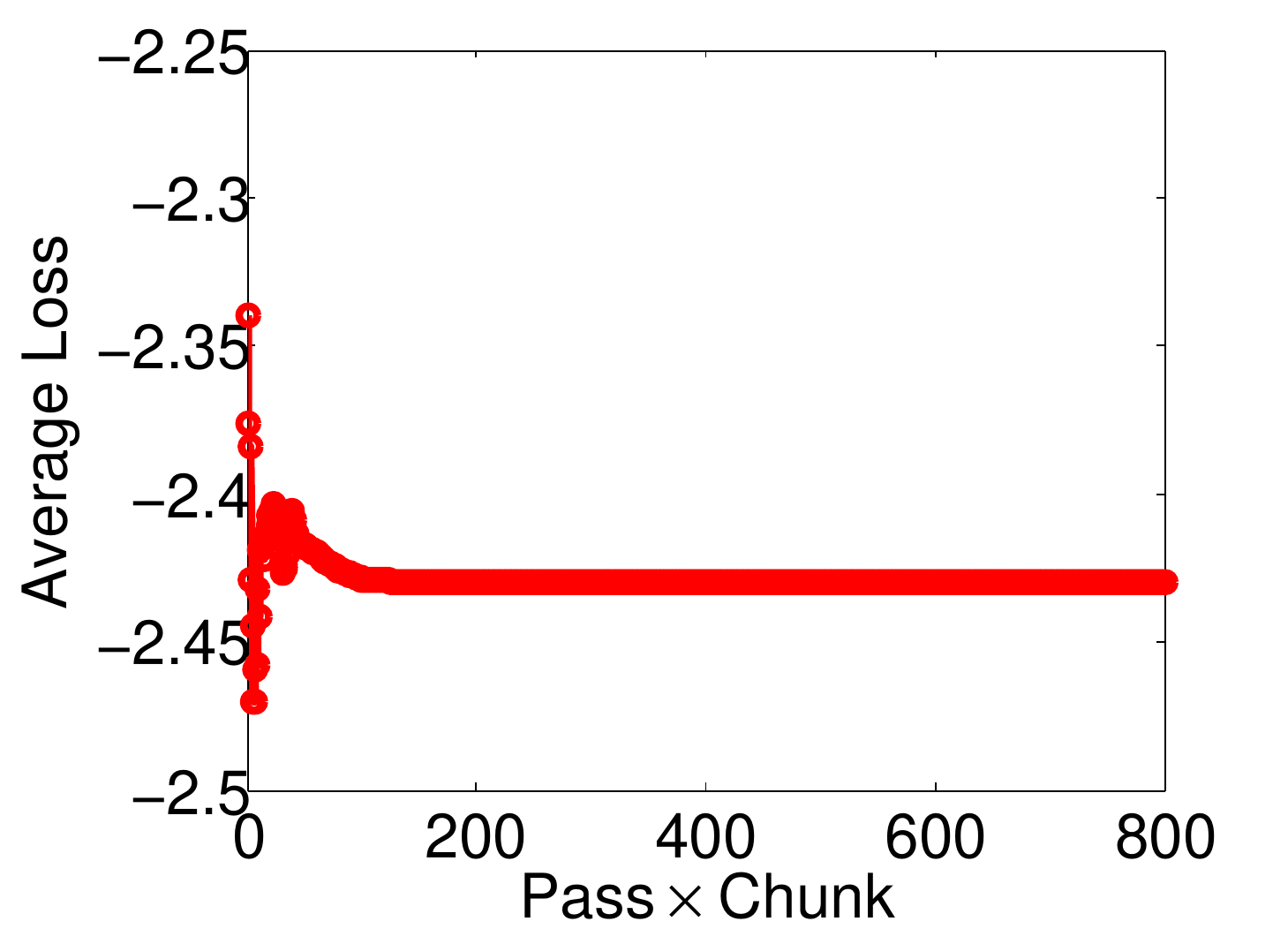}
      }
      \subfigure[Average Loss on Reuters]{
           \label{plot:Reuters}
           \includegraphics[width=3.95cm,height=3cm]{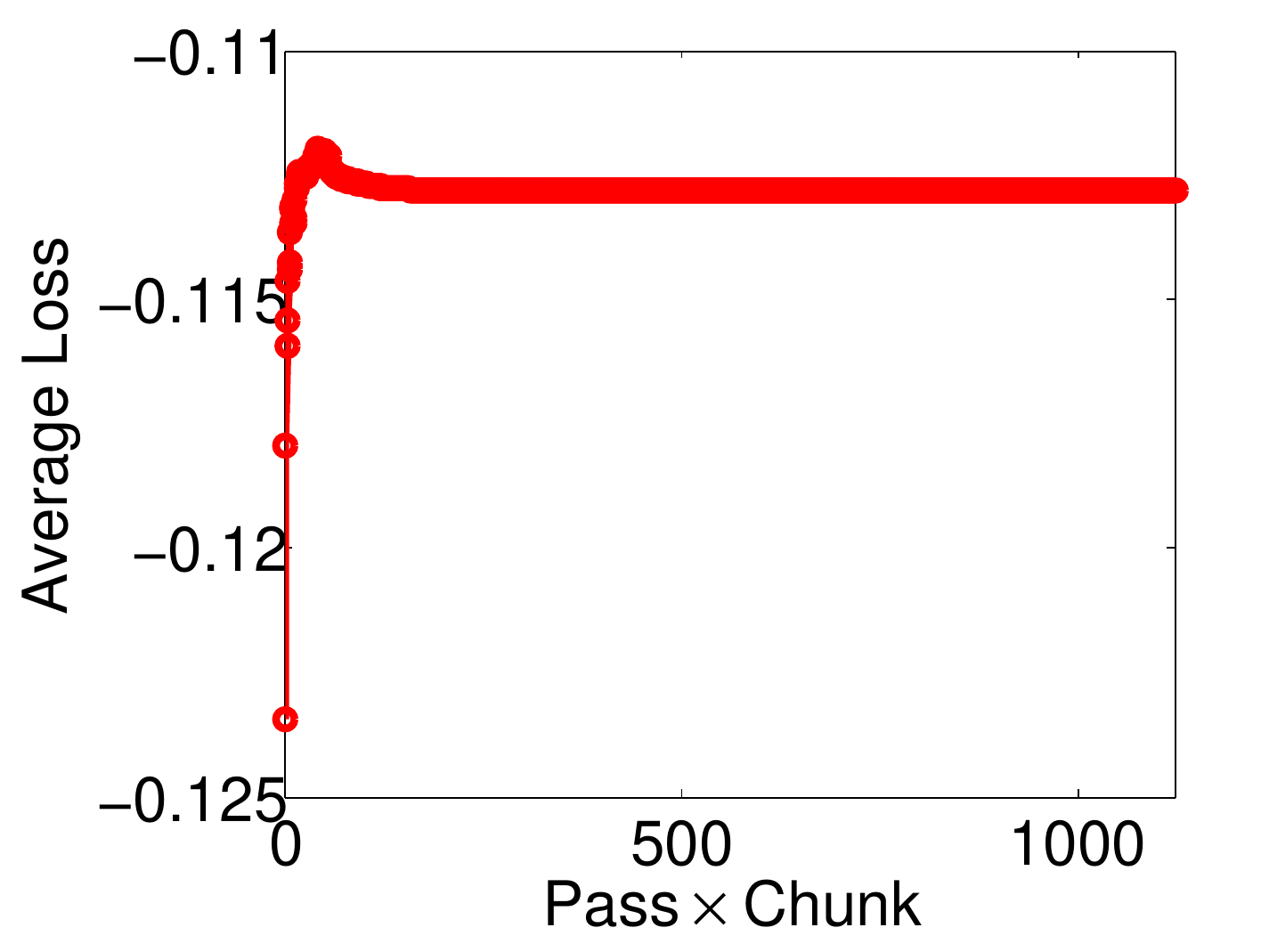}
      }
      \subfigure[Average Loss on Youtube]{
           \label{plot:Youtube}
           \includegraphics[width=3.95cm,height=3cm]{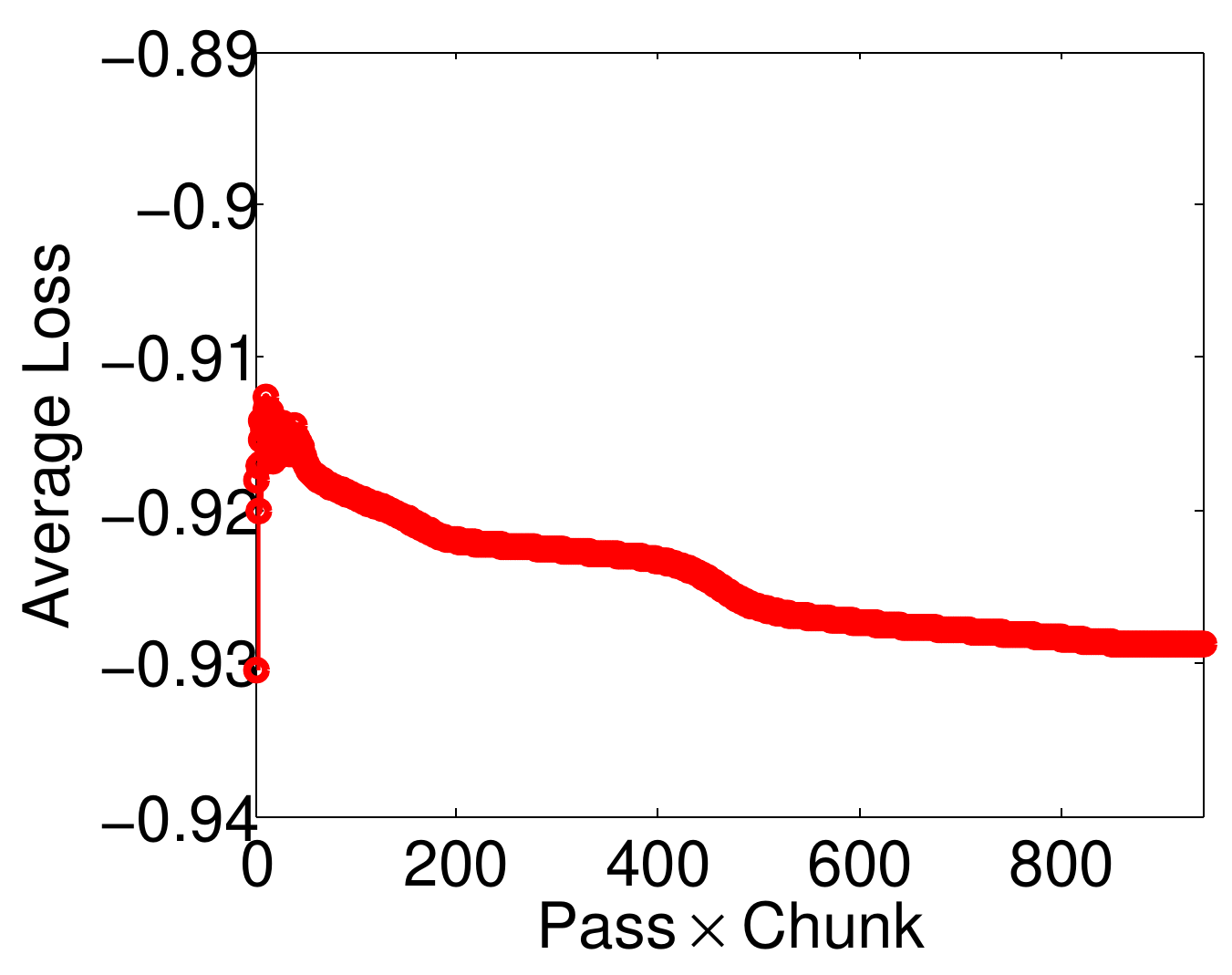}
      }
      \caption{Convergence studies on WebKB, Digit, Reuters and Youtube datasets, where the incomplete rate is set as 0.4, and the experiments are run for 20 passes, the corresponding average loss $\mathcal{L}$ is recorded. It is worth mentioning that since we ignore the loss of $tr(\textbf X^{(v)T}\textbf X^{(v)})$, the average loss $\mathcal{L}$ is negative.}
      \label{Fig:loss}
  \end{figure*}
\begin{equation}
\begin{aligned}
\mathcal{L} = \frac{1}{\min \{s\times t, N\}}\sum_{v=1}^{n_{v}}\left (-2\textbf P_t^{(v)}+\textbf Q_t^{(v)}+\alpha\| \textbf U^{(v)} \|_{\emph{\scriptsize{F}}}^{2}\right )\\
\end{aligned}
\end{equation}
where
\begin{equation}
\begin{aligned}
&\textbf P_t^{(v)} = tr(\textbf U^{(v)T}\textbf R_{t}^{(v)})\\
&\textbf Q_t^{(v)} = tr(\textbf U^{(v)T}\textbf U^{(v)}\textbf T_{t})
\end{aligned}
\end{equation}
And for the other passes, since we can easily count the loss of scanned instances, we define the average loss as follows:
\begin{equation}
\begin{aligned}
\mathcal{L} = \frac{1}{N}\sum_{v=1}^{n_{v}}\left (-2\textbf P_N^{(v)}+\textbf Q_N^{(v)}+\alpha\| \textbf U^{(v)} \|_{\emph{\scriptsize{F}}}^{2}\right )\\
\end{aligned}
\end{equation}

We cascade all pass losses and get the results as shown in Figure \ref{Fig:loss}.

\begin{figure}[htb]
      \centering
      \subfigure[]{
           \label{plot:paraAC}
           \includegraphics[width=3.95cm,height=3.0cm]{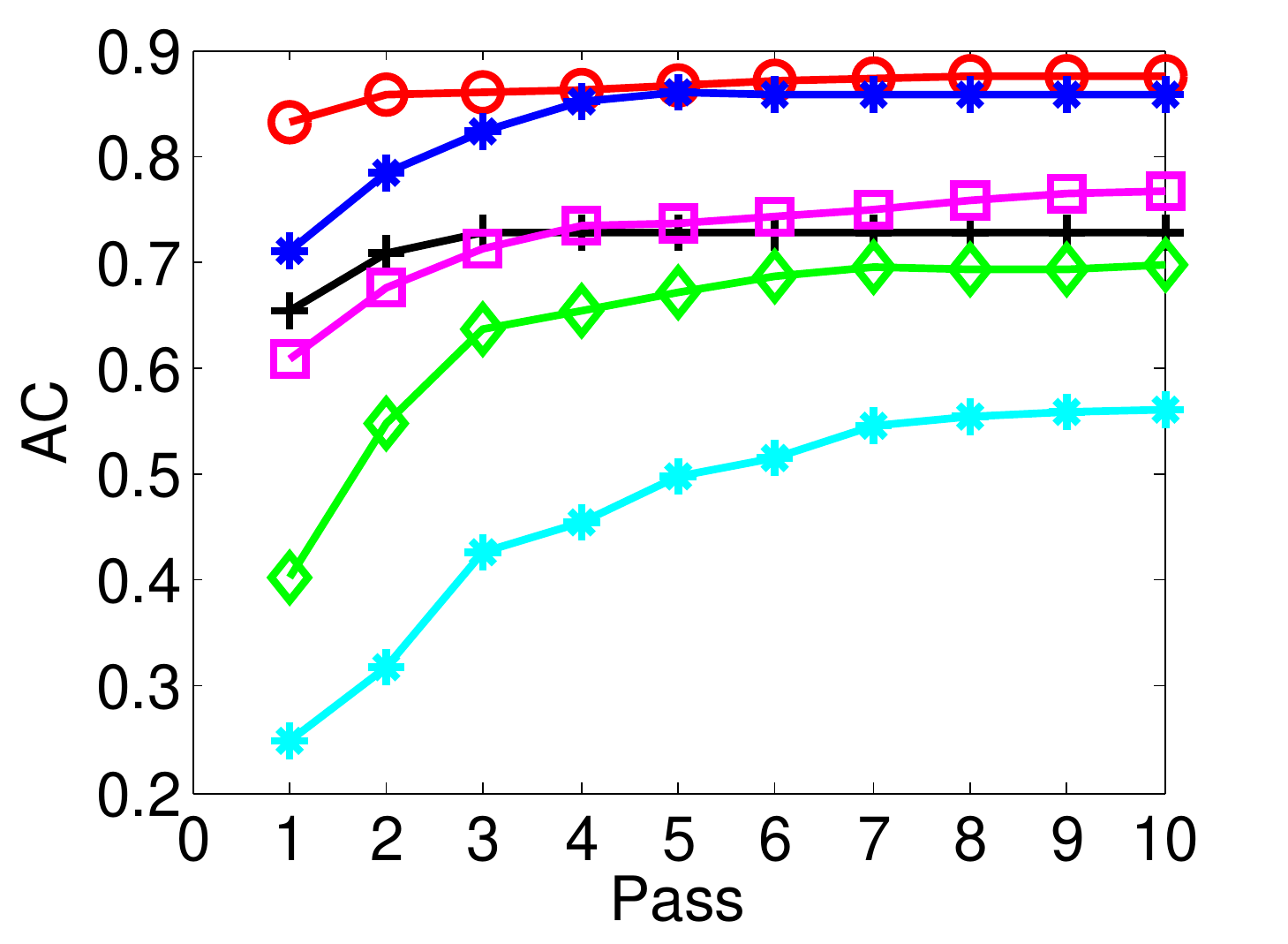}
      }
      \subfigure[]{
           \label{plot:paraNMI}
           \includegraphics[width=3.95cm,height=3.0cm]{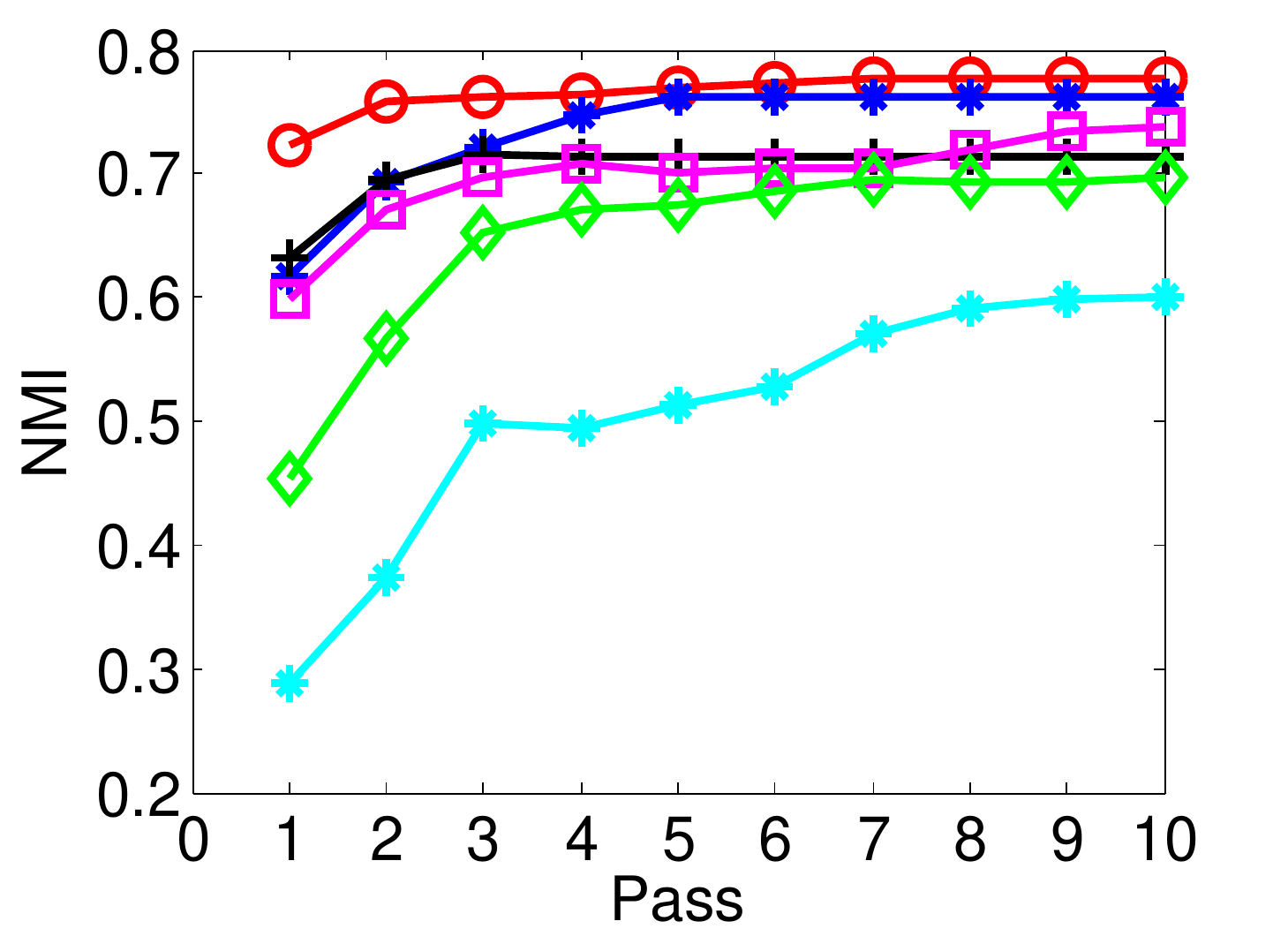}
      }
      \subfigure{
           \label{legend}
           \includegraphics[width=0.4\textwidth]{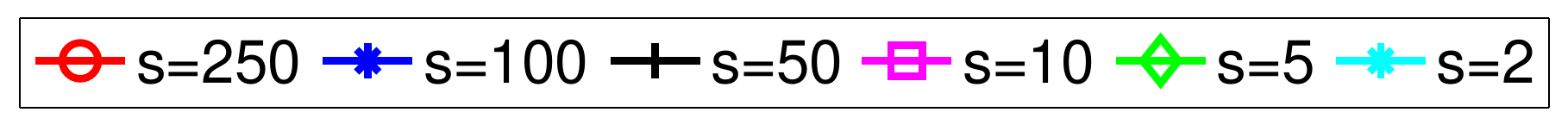}
      }
      \caption{Different block size study on Digit dataset, where the incomplete rate is set as 0.4 and the experiment is run for 10 passes.}
      \label{Fig:block}
  \end{figure}

From Figure \ref{Fig:loss}, we can see that, as the training goes on, the average loss converge gradually. Corresponding to Figure \ref{Fig:NMIAC}, we can observe that when the average loss converges, both NMI and AC get stable values.\\
$\textbf{Block\  Size\ Study}$: In OPIMC, the size of data chunk is a vitally important parameter. In order to study the performance of OPIMC with different chunk sizes, we conduct a block size study on digit dataset. Besides, we set the incomplete rate to 0.4, and report the clustering performance of OPIMC by ranging $s$ in the set of $\{2,5,10,50,100,250\}$. Meanwhile, we run the experiment for 10 passes and the results are shown in Figure \ref{Fig:block}.

\begin{figure}[htb]
      \centering
      \subfigure[]{
           \label{plot:dirftac}
           \includegraphics[width=3.95cm,height=3.0cm]{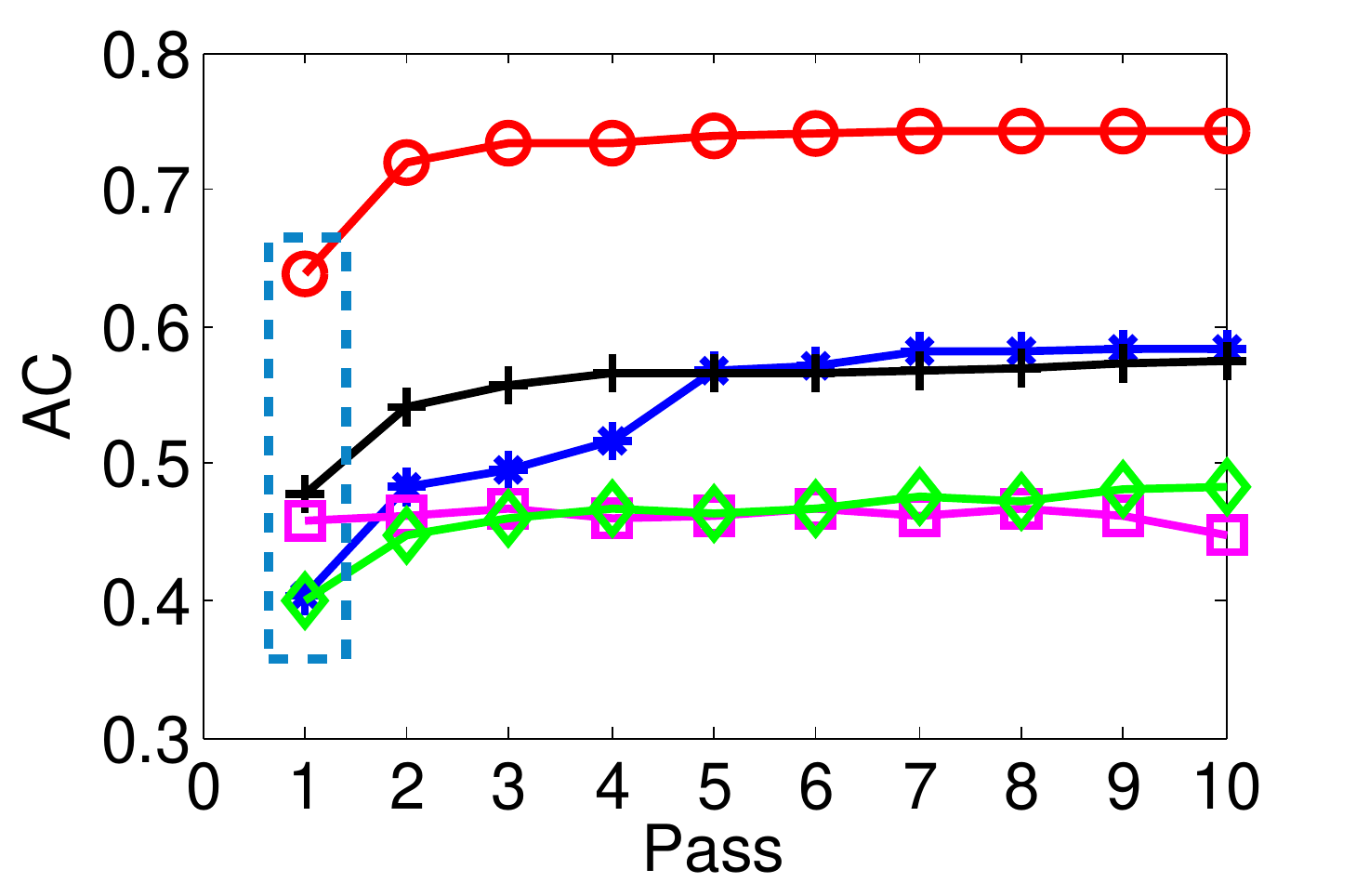}
      }
      \subfigure[]{
           \label{plot:dirftnmi}
           \includegraphics[width=3.95cm,height=3.0cm]{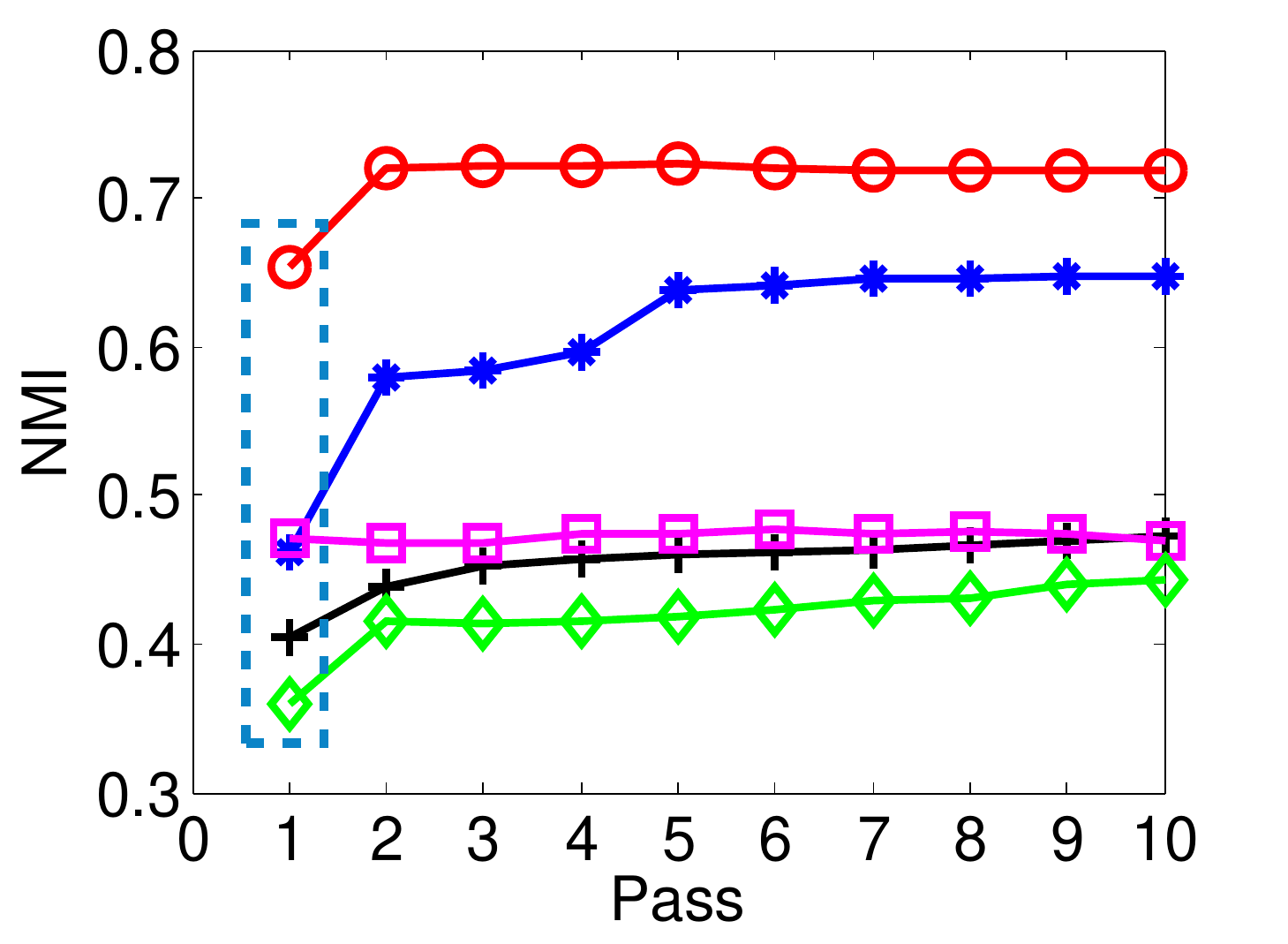}
      }
      \subfigure{
           \label{legend}
           \includegraphics[width=0.4\textwidth]{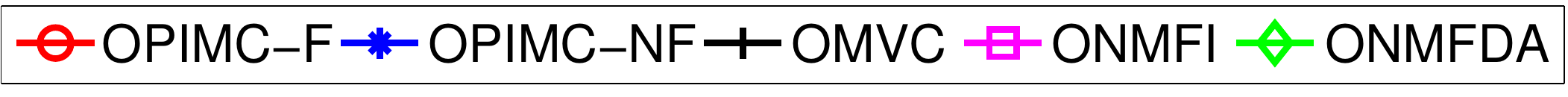}
      }
      \caption{Clustering center degradation study on Digit dataset, where OPIMC-F and OPIMC-NF denote the OPIMC with filled and not filled degraded cluster centers, respectively. Besides, the incomplete rate is set as 0.4 and the experiment is run for 10 passes.}
      \label{Fig:drift}
  \end{figure}

From Figure \ref{Fig:block}, we can see that generally the bigger the block size, the better the clustering results. Furthermore, when $s=250$, the NMI and AC get a great value. However, using larger chunk size will cause larger space complexity.\\
$\textbf{Clustering\ Center\ Degradation\ Study}$: In this experiment, we will prove the validity of filling the degraded cluster centers. we conduct the experiment on Digit dataset with the incomplete rate of 0.4. We do not disrupt the instance order of the Digit dataset and implement OPIMC with filled (OPIMC-F) and not filled (OPIMC-NF) degraded cluster centers, respectively. We run the experiment for 10 passes and the results are shown in Figure \ref{Fig:drift}, from which we can witness the effect of filling degenerate cluster centers very directly.
\section{Conclution}
In this paper, we propose an efficient and effective method to deal with large scale incomplete multi-view clustering problem by adequately considering the instance missing information with the help of regularized matrix factorization and weighted matrix factorization. By introducing two global statistics, OPIMC can directly get clustering results and effectively determine the termination of iteration process. The experimental results on four real-world multi-view datasets demonstrate the efficiency and effectiveness of our method. In the future, the generation of new classes and the robustness of algorithms will be the focus of our consideration.

\newpage
\section{Acknowledgments}
This work is supported in part by the NSFC under Grant No.61672281, and the Key Program of NSFC under Grant No.61732006
\bibliographystyle{aaai}
\bibliography{AAAI-HuM.4562}
\end{document}